\pdfoutput=1
\documentclass[11pt]{article}

\usepackage[final]{acl}
\usepackage{times}
\usepackage{latexsym}

\usepackage[T1]{fontenc}
\usepackage[utf8]{inputenc}

\usepackage{microtype}

\usepackage{inconsolata}

\usepackage{graphicx}

\usepackage{caption}
\usepackage{subcaption}
\usepackage{enumitem}
\usepackage{multirow}
\usepackage{booktabs}
\usepackage{hyperref}
\newcommand{\rom}[1]{\lowercase\expandafter{\romannumeral #1\relax}}

\title{To MRL or not to MRL: Text Embeddings are Robust to Truncation Without Matryoshka Learning, Except In Heavy Truncation Scenarios}

\author{Sotaro Takeshita\textsuperscript{1}, Yurina Takeshita\textsuperscript{3}, Simone Paolo Ponzetto\textsuperscript{1}, Daniel Ruffinelli\textsuperscript{1,2} \\
  \textsuperscript{1}Data and Web Science Group, University of Mannheim, Germany \\
  \textsuperscript{2}NEC Laboratories Europe, Heidelberg, Germany \\
  \textsuperscript{3}Independent Researcher \\ 
  \texttt{\{sotaro.takeshita, ponzetto, druffinelli\}@uni-mannheim.de}}

\usepackage{amsfonts}
\usepackage{amsmath}
\usepackage{bm}
\newcommand\set[1]{\mathcal{#1}}
\newcommand\bb[1]{\mathbb{#1}}
\newcommand\vect[1]{{\boldsymbol{#1}}}

\newcommand\ve{\vect{e}}

\newcommand\vtheta{\vect{\theta}}

\newcommand\mW{\vect{W}}

\begin{document}
\maketitle

\begin{abstract}
Matryoshka Representation Learning (MRL) is a widely adopted approach 
for training text encoders so they provide useful text representations 
at various sizes, available by simply truncating the resulting vectors 
at sizes pre-determined at training time.
Recent works have shown that randomly truncating text embeddings has 
minimal impact in downstream performance unless vectors are reduced in 
size by at least 70\%, suggesting that embeddings are already robust to 
truncation without the use of MRL.
However, no prior work has compared random truncation to MRL, 
so it is unclear how the two 
methods compare as effective embedding reduction methods.
In this paper, we study this by applying the same 
truncation used by MRL to models trained with and without MRL.
Our results across several models and downstream tasks show that,
unless heavily truncating embeddings (i.e.\ reducing their size by at 
least 80\%), truncated embeddings of non-MRL models are
competitive with, and often outperform models trained with MRL.
This suggests that truncation robustness may not necessarily 
come from MRL, and that the choice of spending the additional 
training cost of MRL depends on whether heavy truncation is desired.
We make our code available for reproduction.\footnote{\url{https://sotaro.io/papers/mrl-or-not}}
\end{abstract}

\section{Introduction}\label{sec:intro}

Text embeddings are widely used in many NLP tasks, 
from retrieval~\citep{huang2020embedding} to recommender 
systems~\citep{zhao2023embedding} to many others~\citep{zhao2024dense}.
To reduce costs while retaining performance in such tasks, the use of
small but well-performing embeddings is desirable. 
E.g.\ in text retrieval, bi-encoder models with less than 
\begin{figure}[h!]
    \centering
    \includegraphics[width=0.95\columnwidth]{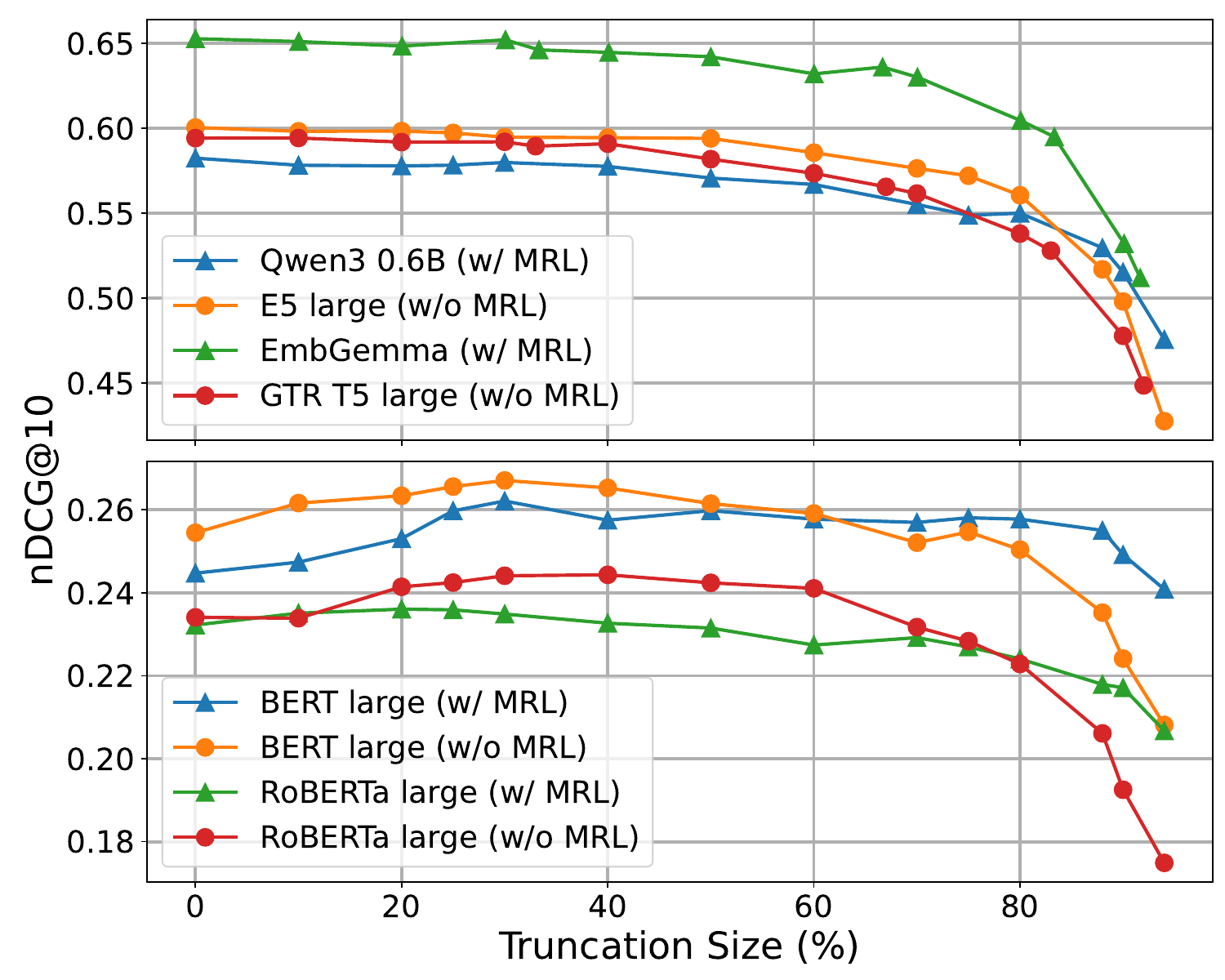}
    \caption{
    (Top) Robustness of open text encoders as truncation levels increase 
    looks the same whether trained with or without MRL.
    (Bottom) When models differ only in their use of MRL,
    truncation on non-MRL models is superior unless heavy truncation is 
    applied. 
    }
    \label{fig:first-page-figure}
\end{figure}
1B parameters are a common choice for first-stage retrieval over an entire 
collection of documents~\citep{zhao2024dense}.
To provide more flexibility in this regard, Matryoshka Representation Learning 
(MRL)~\citep{kusupati2022matryoshka} is an approach that adds additional terms 
to the training objective so that text encoders are able to provide 
representations that are simultaneously useful at various embedding sizes, 
available by simply truncating the resulting vectors at sizes pre-determined 
before training.
The use of MRL is widely adopted by most of the latest models, such as
% , with some of the 
% latest models providing this benefit out-of-the-box, such as 
Jina-Embeddings-V5~\citep{akram2026jina}, Qwen3-Embedding~\citep{zhang2025qwen3embeddingadvancingtext}, 
and EmbeddingGemma~\citep{vera2025embeddinggemma}.
At the same time, recent works have empirically shown that randomly 
truncating embeddings is a promising approach to obtaining vectors 
of reduced size, as the impact in performance is minimal unless vector sizes 
are reduced by at least 
70\%~\citep{tsukagoshi-sasano-2025-redundancy,takeshita2025randomly,inkiriwang2025we}.
However, no prior work has compared random truncation to MRL, the industry 
standard for truncation, so it is unclear to what extent random truncation is 
an effective method for trading off performance for runtime and memory costs.
In this short paper, we study and quantify this by comparing downstream performance 
at various vector truncation levels using embeddings from models trained with and 
without MRL.
Our results show that unless heavily truncating embeddings, i.e.\ reducing 
their size by about 80\%, truncated embeddings of non-MRL models are 
competitive with, and often superior in performance compared to models 
trained with MRL.
We found this across 8 open models (e.g.\ Fig.~\ref{fig:first-page-figure}, top), 10 text encoders 
we trained from pre-trained language models (e.g.\ Fig.~\ref{fig:first-page-figure}, bottom) and 
24 downstream tasks for classification and retrieval.
Upon closer inspection, we found the variance of lower dimensions in 
Matryoshka embeddings to be much larger than non-MRL models, suggesting 
MRL does promote better information compression in these dimensions.
This might relate to why MRL only brings clear benefits in heavy truncation 
settings, i.e.\ in scenarios with potentially more cost reductions, but also 
higher performance loss,
Finally, we found that while MRL requires that truncation dimensions be 
chosen before training, MRL behaves the same as random truncation
when truncating outside of those pre-selected dimensions.
Our results suggest that (i) truncation robustness may not necessarily 
come from MRL but be inherent to the learned representations, 
(ii) that the additional cost of training with MRL is only beneficial in 
scenarios where reducing embedding size by about 80\% or more is desirable, 
and (iii) that there may be ways to extend MRL's benefit beyond heavy truncation 
scenarios.

\section{Background and Related Work}\label{sec:background}

\paragraph{Matryoshka Representation Learning (MRL).}
Following~\citet{kusupati2022matryoshka}, we formalize MRL in a supervised 
setting. Let $f: T\to \bb{R}^d$ be a text encoder with parameters $\vtheta$ that 
maps a sequence of tokens $x\in \set{T}$ to a $d$-dimensional vector $\ve$, 
i.e.\ $\ve = f(x|\vtheta)$, where $\set{T}$ is the set of all possible token
sequences in a given language. 
Given labelled dataset 
$\set{D} = \{(x_1, y_1), (x_2, y_2), \ldots, (x_N, y_N)\}$ and loss function
$\set{L}$, MRL optimizes the following objective:
\begin{align*}
    \min_{\vtheta, \set{W}_{\set{M}}} \frac{1}{N} \sum_{(x,y)\in\set{D}}\sum_{m\in \set{M}} c_m\cdot\set{L}(\mW_{m}\cdot f_m(x|\vtheta), y)
\end{align*}
where $\set{M} = \{m_1, m_2, \ldots, m_{|M|}\}$ is a set of vector 
sizes $m_i\leq d$, $\set{W}_{\set{M}} = \{\mW_{m}\}_{m\in\set{M}}$ the set of
learnable classifiers 
for each $m\in\set{M}$, $c_{m}$ a scaling factor 
for the loss term corresponding to vector size $m$, and 
$f_m(x|\vtheta) = f(x|\vtheta)_{1:m}$ the output vector of encoder $f$ 
truncated to size $m$.
In other words, MRL applies empirical risk minimization over the set of 
classifiers $\set{W}_\set{M}$, where each uses a different truncation of 
vector $\ve$ as input.
Weights are typically shared across all classifiers in $\set{W}_{\set{M}}$
for efficiency, i.e.\ a nested classifier is trained.
Optionally, $\set{W}_{\set{M}}$ can be dropped so that only $\vtheta$
is optimized.

Subsequent works have proposed variants of MRL, e.g.\ 
to inject MRL properties to already trained 
models~\citep{yoon2024matryoshka}, or to 
reduce the number of layers in a model~\citep{2d2024wang}. 
More recently, \citet{zhang2025smec} compared MRL other forms of size reduction. 
However, no such studies have used random truncation as a baseline.

\paragraph{Random truncation.}

Inspired by the fact that the proof in the seminal Johnson-Lindenstrauss 
lemma~\citep{johnson1984extensions} is based on random projections, many
works have since based their embedding-based methods on random 
projections~\citep{achlioptas2003database}. 
More recently, \citet{takeshita2025randomly} found that randomly 
truncating text embeddings in various ways results in minimal
performance loss unless vectors are heavily reduced in size.
Similarly, \citet{tsukagoshi-sasano-2025-redundancy} also reported
the same efficacy of using random truncation as a means of dimensionality
reduction.

Closest to our work, \citet{inkiriwang2025we} compared the efficacy of 
several methods for dimensionality reduction, including random
projections, which they found to be the most consistently effective.
However, while they included more dimensionality reduction methods, we 
include more recent models and a larger variety of downstream tasks. 
More importantly, they do not include MRL in their comparison, despite it 
being used by some of the best performing text encoders to 
date~\citep{sturua2024jina,zhang2025qwen3embeddingadvancingtext,vera2025embeddinggemma}.

\section{MRL vs non-MRL}\label{sec:mrl_vs_random}

\subsection{Benchmarking Open Encoders}\label{sec:open-models}

\begin{figure}[ht]
    \centering
    \begin{subfigure}[b]{0.95\columnwidth}
      \includegraphics[width=0.95\columnwidth]{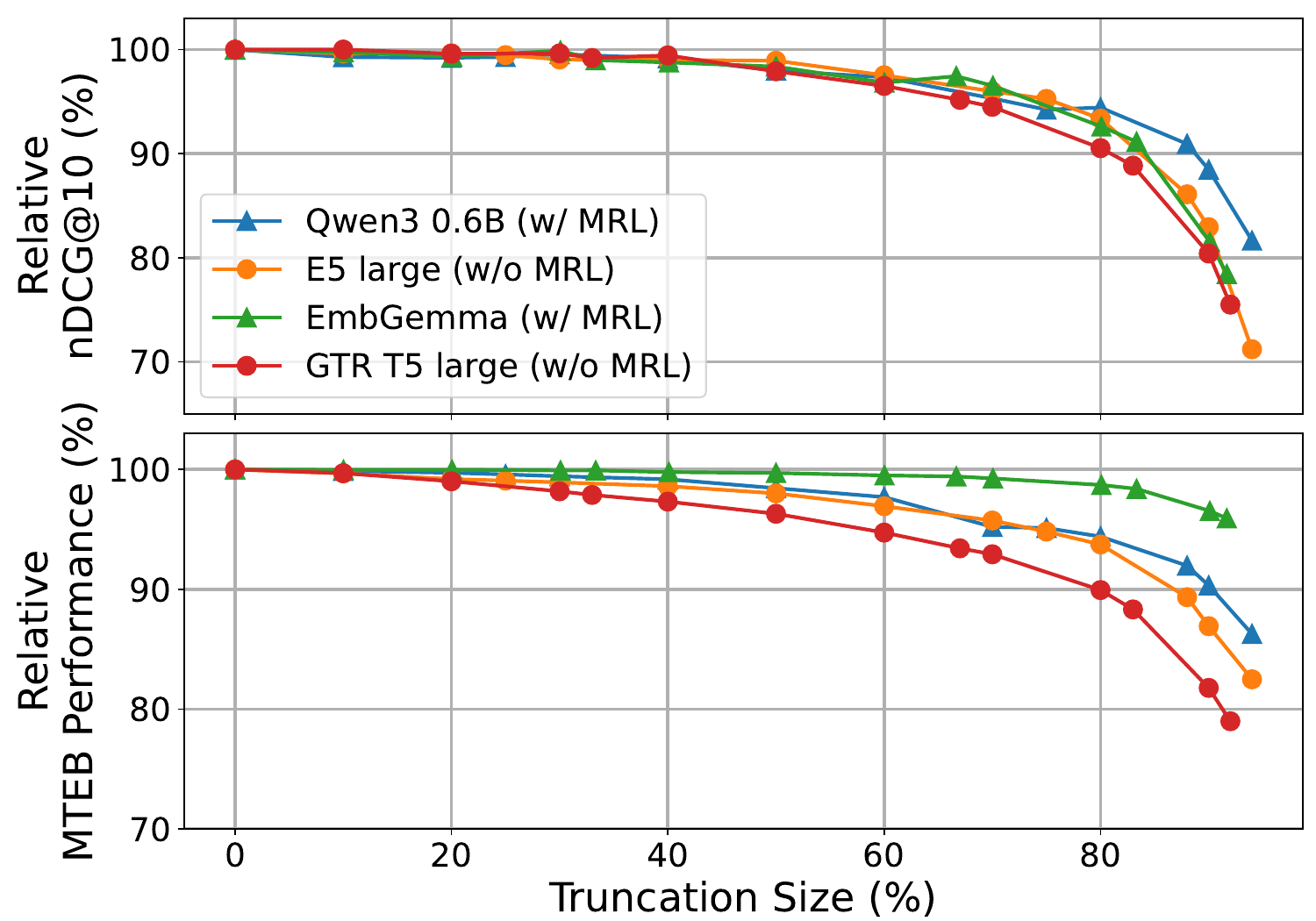}
      \caption{Smaller models ($\sim$400M parameters)}
      \label{fig:relative_performance_smaller_models}
    \end{subfigure}
    % \medskip % insert a bit of vertical whitespace
    \begin{subfigure}[b]{0.95\columnwidth}
      \includegraphics[width=0.95\columnwidth]{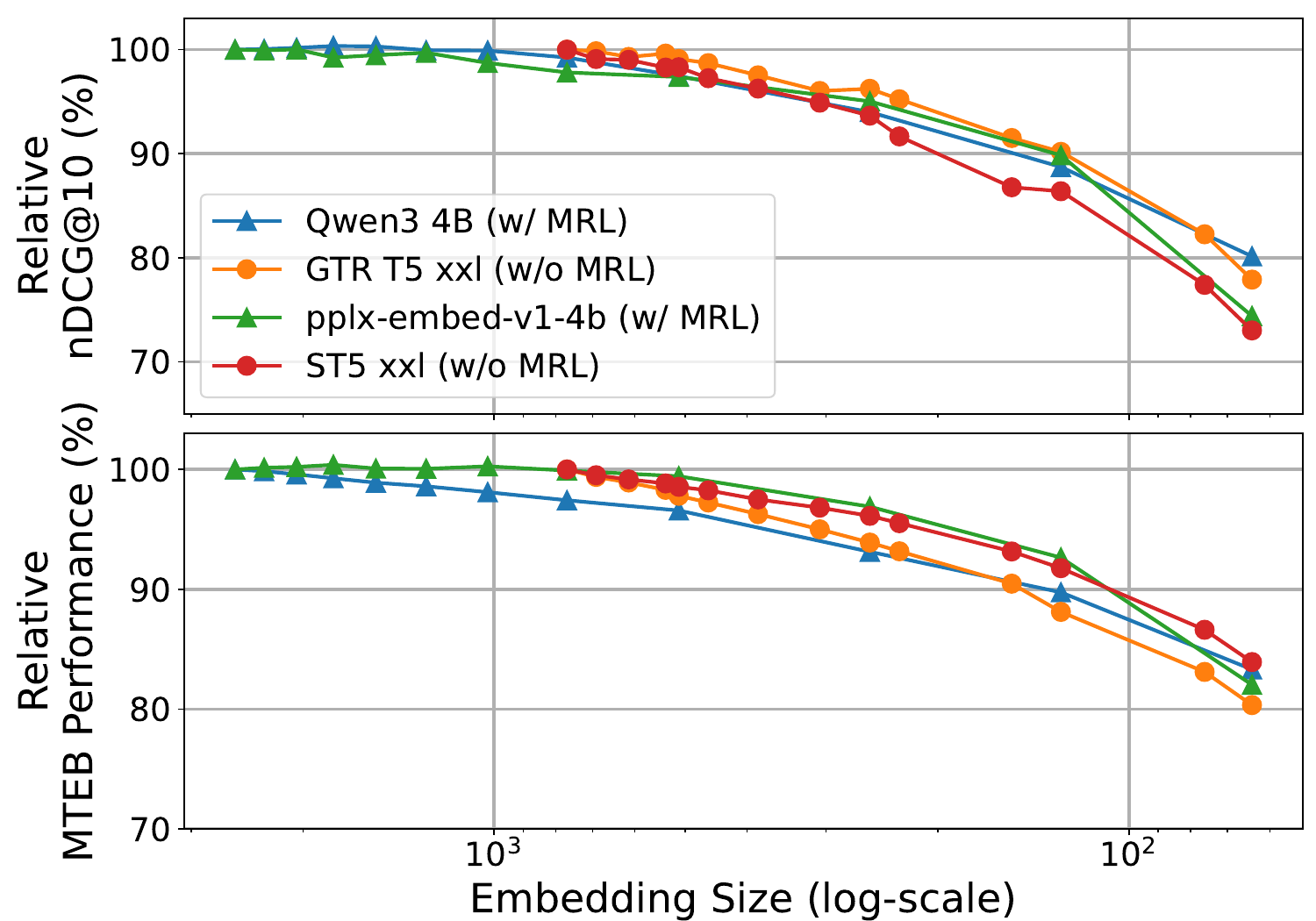}
      \caption{Larger models ($\sim$4B parameters)}
      \label{fig:relative_performance_larger_models}
    \end{subfigure}
    \caption{
        Performance on NanoBEIR (top) and MTEB (bottom) of text embeddings truncated at 
        various sizes, relative to the performance of the corresponding full-size embeddings.
    }
    \label{fig:relative_performance_all_models}
\end{figure}

In this section, we compare publicly available text encoders trained with and without MRL.
Note that while there are many more differences between these models (e.g. architecture, 
training recipe), our goal is to study the impact that MRL has on
truncation robustness in seminal and state-of-the-art models.
It is common to compare text encoders across different architectures,
e.g.\ \citet{wang-etal-2024-improving-text,tsukagoshi-sasano-2025-redundancy,vera2025embeddinggemma},
as other aspects typically differentiate new models from prior work, 
e.g.\ training recipe~\citep{neelakantan2022text,sturua2024jina}. 
This makes a proper comparison prohibitely expensive.
However, we do conduct a more controlled experiment in Sec.~\ref{sec:trained-models}
using a common training recipe.
Following MRL, we use prefix truncation as our truncation strategy on all models,
that is, we remove the last elements of the embeddings.
This means our form of truncation is indeed deterministic.
We opt for prefix truncation among other possible strategies
because an existing study by~\citet{takeshita2025randomly} show
that different truncation strategies do not lead to
noticeable performance differences for non-MRL models.

\paragraph{Models.}
We used eight seminal/state-of-the-art open-weight models, 
four smaller ones ($\sim$500M params.) and four larger ones ($\sim$4B params.).
For non-MRL models, we used E5-Large~\citep{wang-etal-2024-improving-text}, 
GTR-T5-Large and -XXL~\citep{ni-etal-2022-large}, and ST5-XXL~\citep{ni-etal-2022-sentence}.
For MRL, we used Qwen3-Embedding-0.6B and -4B~\citep{zhang2025qwen3embeddingadvancingtext}, 
EmbeddingGemma-300M~\citep{vera2025embeddinggemma}, and 
pplx-embed-v1-4b~\citep{eslami2026diffusionpretraineddensecontextualembeddings}.
For model details, see Appendix~\ref{sec:appendix_experimental_details}.

\paragraph{Evaluation.}
For retrieval, we use 13 datasets from NanoBEIR,
a smaller variant of BEIR~\citep{thakur2021beir}, commonly used for 
research~\citep{conti2025context,vujanic2026leafknowledgedistillationtext}.
For classification, we use 11 datasets from MTEB~\citep{muennighoff-etal-2023-mteb}. 
We list all datasets in Appendix~\ref{sec:appendix_experimental_details}
and report average performance across tasks
(see per-dataset results in Appendix~\ref{sec:appendix_performance_open_per_dataset}).
We report both relative performance 
of truncated embeddings compared to the original full-sized embeddings, 
and absolute performance (in Appendix~\ref{sec:appendix_performance_open_models}). 
We evaluate at embedding sizes that match common MRL training choices: 
64, 128, 256, 512, and 768.
We also evaluate at relative positions of 10\% to 90\% (step 10\%) 
of the original embedding size, which result in sizes outside pre-selected 
MRL sizes.

\paragraph{Results and discussion.}
Looking at Fig.~\ref{fig:relative_performance_all_models},
if MRL were to bring an advantage, we should see that performance 
of MRL models drops more slowly than non-MRL models.
However, we see no clear influence from MRL, as non-MRL models show similar levels
of robustness to truncation, especially in larger models where we see no difference
between MRL and non-MRL models at all truncation levels (see 
Fig.~\ref{fig:relative_performance_larger_models}).
This is true for both the pre-selected MRL sizes and those outside this choice.
The exception is EmbeddingGemma on MTEB, especially in contrast to GTR-T5 which 
drops faster than other models, but this exception does not hold for NanoBEIR 
(see Fig.~\ref{fig:relative_performance_smaller_models}).
And performance in absolute terms does not seem to provide an explanation
(see Fig.~\ref{fig:open-small-beir-mteb-absolute} in Appendix~\ref{sec:appendix_performance_open_models}), as EmbeddingGemma consistently 
outperforms all models, but is more robust to truncation only in MTEB and not NanoBEIR, 
where Qwen is more robust beyond 80\% truncation, despite underperforming 
in absolute terms.
Together, these results suggest that MRL does not provide higher robustness to
truncation, except at very high truncation levels and mostly on smaller
models. 
However, while it is important to evaluate the impact of MRL on state-of-the-art 
models, it is difficult to reason about these results given that any of the
various differences between models, and not just MRL, may account for the 
observed results.
So, we report results from a more controlled experiment in the next section.

\subsection{Benchmarking Trained Encoders}\label{sec:trained-models}

\begin{figure}[t]
    \centering
    \includegraphics[width=0.95\columnwidth]{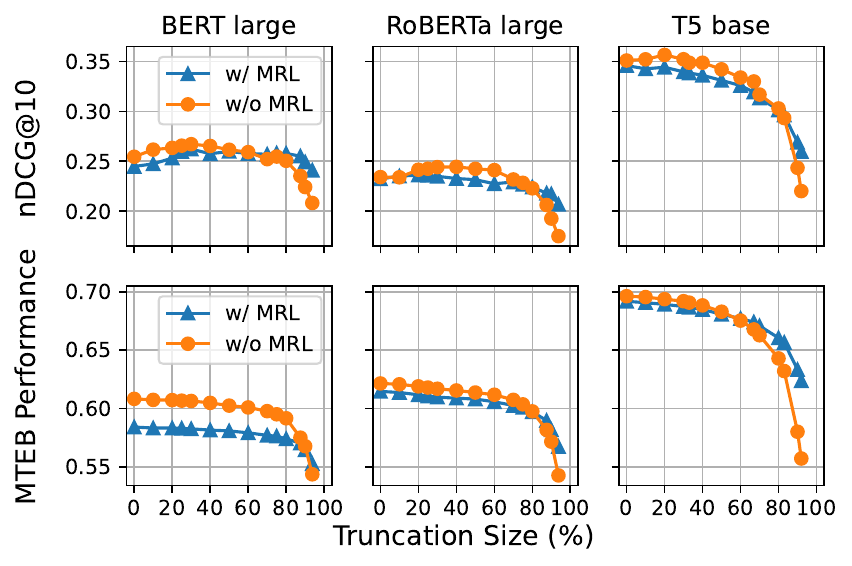}
    \caption{
        Performance on BEIR and MTEB benchmarks of three pairs of encoders 
        trained with and without MRL. 
        See all models in Fig.~\ref{fig:trained-models-all} in 
        Appendix~\ref{sec:appendix_performance_trained_models}.
    }
    \label{fig:trained-models}
\end{figure}

To better determine the impact of MRL on embedding truncation,
in this section we train pairs of encoders where the only difference 
is the use of MRL during training.
Unless otherwise noted, experimental settings are the same as in
Sec.~\ref{sec:open-models}.

\paragraph{Models.}
We take five pre-trained language models (PLMs) from different sizes and 
architectures as starting points for contrastive training.
We train two encoder-only models: BERT~\citep{devlin-etal-2019-bert} and RoBERTa~\citep{liu2020roberta},
each in two sizes (base and large).
We also train the encoder of T5~\citep{t5}, an encoder-decoder model.

\paragraph{Contrastive training.}
We train all PLMs with multiple negative ranking loss~\citep{henderson2017efficient} 
both for non-MRL and MRL models.
For training and validation data, we use the concatenation of 
SNLI~\citep{bowman-etal-2015-large} and MultiNLI~\citep{williams-etal-2018-broad}.
These choices of loss function and training data are common in prior
work~\citep{vulic-etal-2023-probing,ni-etal-2022-sentence,gao-etal-2021-simcse}.
For MRL models, we uniformly weigh the losses for each of the
following nested dimensions: 64, 128, 256, 512 and 768.
We train models until convergence (see Fig.~\ref{fig:training-log-all-models} in Appendix~\ref{sec:appendix_convergence}) 
and select the best models on held-out validation samples.

\paragraph{Results and discussion.}
As before, we evaluate models on MTEB and 
BEIR.\footnote{BEIR for smaller models (BERT base and RoBERTa base) and NanoBEIR for the rest.} 
Fig.~\ref{fig:trained-models} shows how average downstream performance changes 
with different levels of truncation for MRL and non-MRL models (see 
Appendix~\ref{sec:appendix_performance_trained_per_dataset} for results per dataset).
Surprisingly, truncated embeddings from non-MRL models outperform their 
MRL counterparts almost every time across all models and datasets up until 
about when 80\% truncation is applied, at this crossover point MRL truncation becomes 
increasingly more beneficial (see Appendix~\ref{sec:crossover} on how we compute the crossover point).
The slight drop in performance in MRL models may be due to the additional constraints 
in the training objective, but more experiments are needed to understand this.
In other words, in a controlled environment, we see that truncated vectors
are more effective without MRL than with MRL, unless heavy truncation is applied.
As with the results in the previous section, these results show that prefix truncation 
is often a more effective method of cost reduction than MRL, as it does not rely on 
any additional training cost and often outperforms truncation done on MRL models.
However, MRL does become beneficial at heavy truncation levels, where there is more
potential for cost reductions but also more potential for performance loss.
To better understand this, we plotted the standard deviation across embedding features 
in our trained PLMs using queries from NanoBEIR's MS Marco as input. 
We found that the variance is much higher in lower dimensions 
with MRL models (see Fig.~\ref{fig:std-bert-base-uncased} for 
BERT, Fig.~\ref{fig:std-all-models} in Appendix~\ref{sec:appendix_std_per_dimension} 
for all models). 
This suggests that MRL indeed induces more information storage in the
lower dimensions of text embeddings. However, it is unclear why this is not
the case for all dimensions, or whether MRL can be improved to
achieve this at all levels of truncation.

\begin{figure}[t]
    \centering
    \includegraphics[width=0.95\columnwidth]{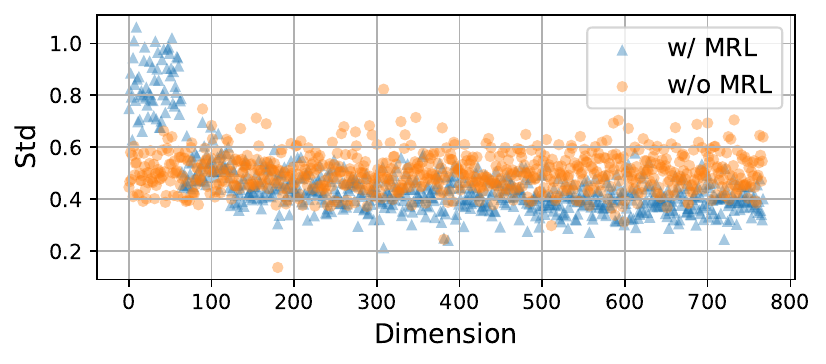}
    \caption{Standard deviation across embedding dimensions from BERT base
    for different text inputs.
    MRL increases the variance of the lower dimensions.
    }
    \label{fig:std-bert-base-uncased}
\end{figure}

\section{Conclusion}
We studied robustness to truncation of text embeddings obtained 
with and without Matryoshka Representation Learning (MRL), a 
training method designed for effective truncation.
Our test with 8 open text encoders, 10 newly trained encoders and 
24 downstream tasks from classification and retrieval benchmarks,
show that MRL provides clear benefits at very 
high truncation levels (higher than 80\%), but that at lower levels 
is not more beneficial than applying truncation on models without MRL
(the precise point is model- and task-dependent).
This suggests future work could improve MRL to provide robustness at 
all truncation levels.

\section*{Limitations}
This work has several limitations.
First, while the training dataset used for our contrastive training 
is a common choice in prior work, there are more recent training 
recipes with larger datasets to train more powerful encoders. 
We opted for our current choice due to limited computational resources.
Second, we only trained small text encoders, again due to limited 
computational resources. It may be that repeating our experiments from 
Section~\ref{sec:trained-models} with larger models produces different
results, and that MRL has a better impact in such models, though results
in Sec.~\ref{sec:open-models} do not suggest this.
Third, a larger variety of settings for MRL training could be tested, e.g. 
exploring the impact that different choices of MRL's pre-determined truncation
sizes have on performance.
Finally, while our experiments are based on two common application scenarios
of text embeddings, namely classification and retrieval, the real world
applications are not limited to these two such as clustering or retrieval
augmented generation (RAG).
Despite these limitations, we believe we have provided enough empirical
evidence to suggest that MRL needs further attention so that we may 
improve embedding robustness to truncation.

% Bibliography entries for the entire Anthology, followed by custom entries
%\bibliography{anthology,custom}
% Custom bibliography entries only
\bibliography{references}

\appendix
\newpage
\onecolumn
\section{Additional Experimental Details}\label{sec:appendix_experimental_details}
\begin{table*}[h!]
    \footnotesize
    \centering
    \setlength{\tabcolsep}{1.5pt}
    \begin{tabular}{llll}
      \toprule
      
       & \multicolumn{1}{c}{\textbf{Name}} & \multicolumn{1}{c}{\textbf{Domain}} &  \multicolumn{1}{c}{\textbf{Licence}} \\
      \midrule
      \multirow{14}{*}{\rotatebox[origin=c]{90}{\textbf{Retrieval}}}& \href{https://microsoft.github.io/msmarco/}{MS MARCO}~\citep{nguyen2016ms} & Misc. & MIT \\
      & \href{https://www.cl.uni-heidelberg.de/statnlpgroup/nfcorpus/}{NFCorpus}~\citep{boteva2016full} & Bio-Medical & N/A\\
      & \href{https://sites.google.com/view/fiqa/}{FiQA-2018}~\citep{fiqa} & Finance & N/A \\
      & \href{https://huggingface.co/datasets/BeIR/arguana}{ArguAna}~\citep{wachsmuth-etal-2018-retrieval} & Misc. & CC BY 4.0 \\
      & \href{https://webis.de/events/touche-20/}{Touche-2020}~\citep{bondarenko2020overview} & Misc. & CC BY 4.0 \\
      & \href{https://www.quora.com/q/quoradata/First-Quora-Dataset-Release-Question-Pairs}{Quora}& Quora & N/A \\
      & \href{https://github.com/iai-group/DBpedia-Entity/}{DBPedia}~\citep{dbpedia} & Wikipedia & CC BY-SA 3.0 \\
      & \href{https://allenai.org/data/scidocs}{SCIDOCS}~\citep{cohan-etal-2020-specter} & Scientific & GNU General Public License v3.0 \\
      & \href{http://fever.ai/}{FEVER}~\citep{thorne-etal-2018-fever} & Wikipedia & CC BY-SA 3.0 l \\
      & \href{http://climatefever.ai/}{Climate-FEVER}~\citep{leippold2020climatefever} & Wikipedia & N/A \\
      & \href{https://github.com/allenai/scifact}{SciFact}~\citep{wadden-etal-2020-fact} & Scientific & CC BY-NC 2.0 \\
      & \href{https://github.com/google-research-datasets/natural-questions}{Natural Questions}~\citep{kwiatkowski-etal-2019-natural} & Scientific & CC BY-SA 3.0 \\
      & \href{https://hotpotqa.github.io/}{HotpotQA}~\citep{yang-etal-2018-hotpotqa} & Scientific & CC BY-SA 4.0 \\
      & \href{https://ir.nist.gov/covidSubmit/index.html}{TREC-COVID}~\citep{TREC-COVID} & Bio-Medical & Dataset License Agreement \\ 
      \midrule

      \multirow{12}{*}{\rotatebox[origin=c]{90}{\textbf{Classification}}}& \href{https://huggingface.co/datasets/mteb/amazon_counterfactual}{AmazonCounterfactualClassification}~\citep{oneill-etal-2021-wish} & Reviews, Written & CC-by-4.0 \\
      & \href{https://huggingface.co/datasets/mteb/amazon_polarity}{AmazonPolarityClassification}~\citep{10.1145/2507157.2507163} & Reviews, Written & Apache 2.0 \\
      & \href{https://huggingface.co/datasets/mteb/AmazonReviewsClassification}{AmazonReviewsClassification}~\citep{keung-etal-2020-multilingual} & Reviews, Written & N/A \\
      & \href{https://huggingface.co/datasets/mteb/banking77}{Banking77Classification}~\citep{casanueva-etal-2020-efficient} & Written & MIT \\
      & \href{https://huggingface.co/datasets/mteb/emotion}{EmotionClassification}~\citep{saravia-etal-2018-carer} & Social, Written & N/A \\
      & \href{https://huggingface.co/datasets/mteb/imdb}{ImdbClassification}~\citep{maas-etal-2011-learning} & Reviews, Written & N/A \\
      & \href{https://huggingface.co/datasets/mteb/amazon_massive_intent}{MassiveIntentClassification}~\citep{fitzgerald-etal-2023-massive} & Spoken & Apache 2.0 \\
      & \href{https://huggingface.co/datasets/mteb/amazon_massive_scenario}{MassiveScenarioClassification}~\citep{fitzgerald-etal-2023-massive} & Spoken & Apache 2.0 \\
      & \href{https://huggingface.co/datasets/mteb/mtop_domain}{MTOPDomainClassification}~\citep{li-etal-2021-mtop} & Spoken & N/A \\
      & \href{https://huggingface.co/datasets/mteb/mtop_intent}{MTOPIntentClassification}~\citep{li-etal-2021-mtop} & Spoken & N/A \\
      & \href{https://huggingface.co/datasets/mteb/tweet_sentiment_extraction}{TweetSentimentExtractionClassification}~\citep{tweet-sentiment-extraction} & Social, Written & N/A \\

      \bottomrule
      
    \end{tabular}
    \caption{
    A list of datasets used in our evaluation. TREC-COVID is only available in BEIR, not in 
    NanoBEIR~(\url{https://huggingface.co/collections/zeta-alpha-ai/nanobeir}).
    }
    \label{tab:datasets}
\end{table*}

\paragraph{Details about non-MRL models.}
\href{https://huggingface.co/sentence-transformers/sentence-t5-xxl}{ST5 xxl}~\citep{ni-etal-2022-sentence} is an encoder with 5B parameters derived from T5~\citep{t5}. 
While the original T5 is built on an encoder-decoder architecture, the ST5 series takes only the encoder and trains it with a contrastive learning objective.
\href{https://huggingface.co/sentence-transformers/gtr-t5-xxl}{GTR xxl} (5B parameters) and \href{https://huggingface.co/sentence-transformers/gtr-t5-large}{GTR T5 large} (300M parameters)~\citep{ni-etal-2022-large} are also derived from encoders of T5.
The training recipe focuses more on retrieval tasks, e.g., the training data includes datasets such as MS MARCO~\citep{nguyen2016ms} while ST5 is more generic for various embedding tasks.
Both ST5 and GTR are trained without MRL objectives.
\href{https://huggingface.co/intfloat/multilingual-e5-large}{E5 large} (300M parameters)~\citep{wang2024multilingual} is a non-MRL encoder derived from a BERT model.
In addition to the mix of public data, the authors used the text pairs synthetically generated by proprietary language models for their contrastive training.

\paragraph{Details about MRL models.}
\href{https://huggingface.co/google/embeddinggemma-300m}{EmbeddingGemma} (300M parameters)~\citep{vera2025embeddinggemma} is an encoder model derived from Gemma, a decoder-only model. To achieve the architectural conversion, the authors first turned the decoder-only model into an encoder-decoder architecture through a method proposed by~\citet{zhang2025encoderdecodergemmaimprovingqualityefficiency}.
Second, similarly to ST5, they took only the encoder and trained it contrastively to obtain the final encoder model.
EmbeddingGemma provides the original embedding with a size of 768, and is trained with MRL to produce 512, 256, and 128-dimensional embeddings.
\href{https://huggingface.co/Qwen/Qwen3-Embedding-4B}{Qwen3 Embedding 4B} and \href{https://huggingface.co/Qwen/Qwen3-Embedding-0.6B}{Qwen3 Embedding 0.6B}~\citep{zhang2025qwen3embeddingadvancingtext} are built on the Qwen3 text generation model family through multi-stage training and model merging.
The authors trained the models to support more than 100 languages and cover various domains such as code retrieval and bitext mining.
The original embedding dimensions are 2560 and 1024, respectively.
Both models are trained with MRL with the smallest dimension of 32.
The other supported dimensions are not reported in the corresponding documentation and technical report.
\href{https://huggingface.co/perplexity-ai/pplx-embed-v1-4b}{pplx-embed-v1-4b}~\citep{eslami2026diffusionpretraineddensecontextualembeddings} is also trained from Qwen3 text generation model however by a different organization. While this variant has also gone through multi-stage training and model merging, the authors apply a diffusion training method~\citep{gong2025scaling} to enable bidirectional text encoding.
This model is trained with MRL.

\section{Validation Loss Curve During Contrastive Trainings}\label{sec:appendix_convergence}
\begin{figure}[h!]
    \centering
    \includegraphics[width=0.8\columnwidth]{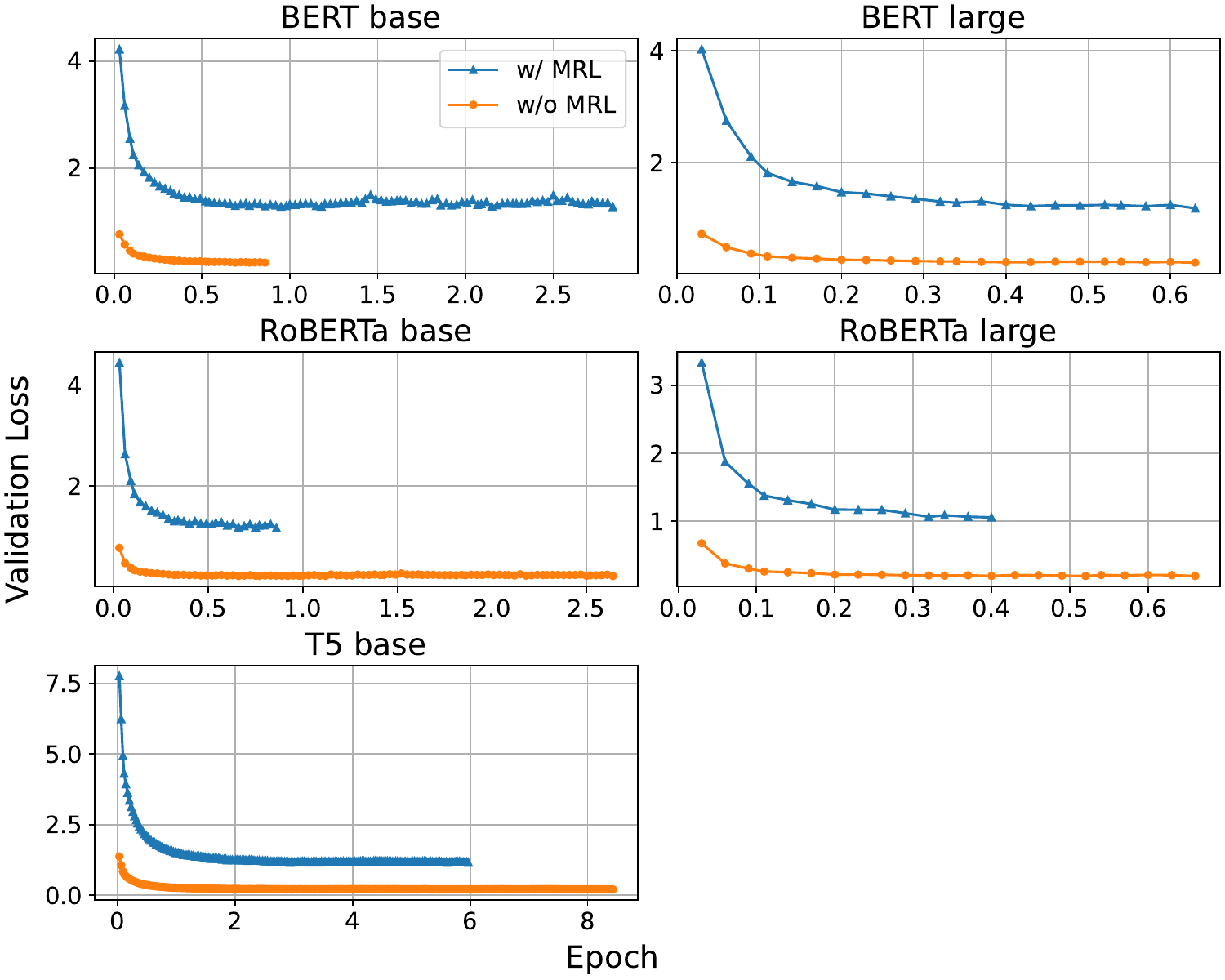}
    \caption{
        Validation loss curve for contrastive learning with and without MRL for all model pairs.
        Our training recipe allowed all models to reach convergence.
    }
    \label{fig:training-log-all-models}
\end{figure}

\newpage
\section{Absolute Performance of Open Models}\label{sec:appendix_performance_open_models}
\begin{figure}[h!]
    \centering
    \includegraphics[width=0.6\columnwidth]{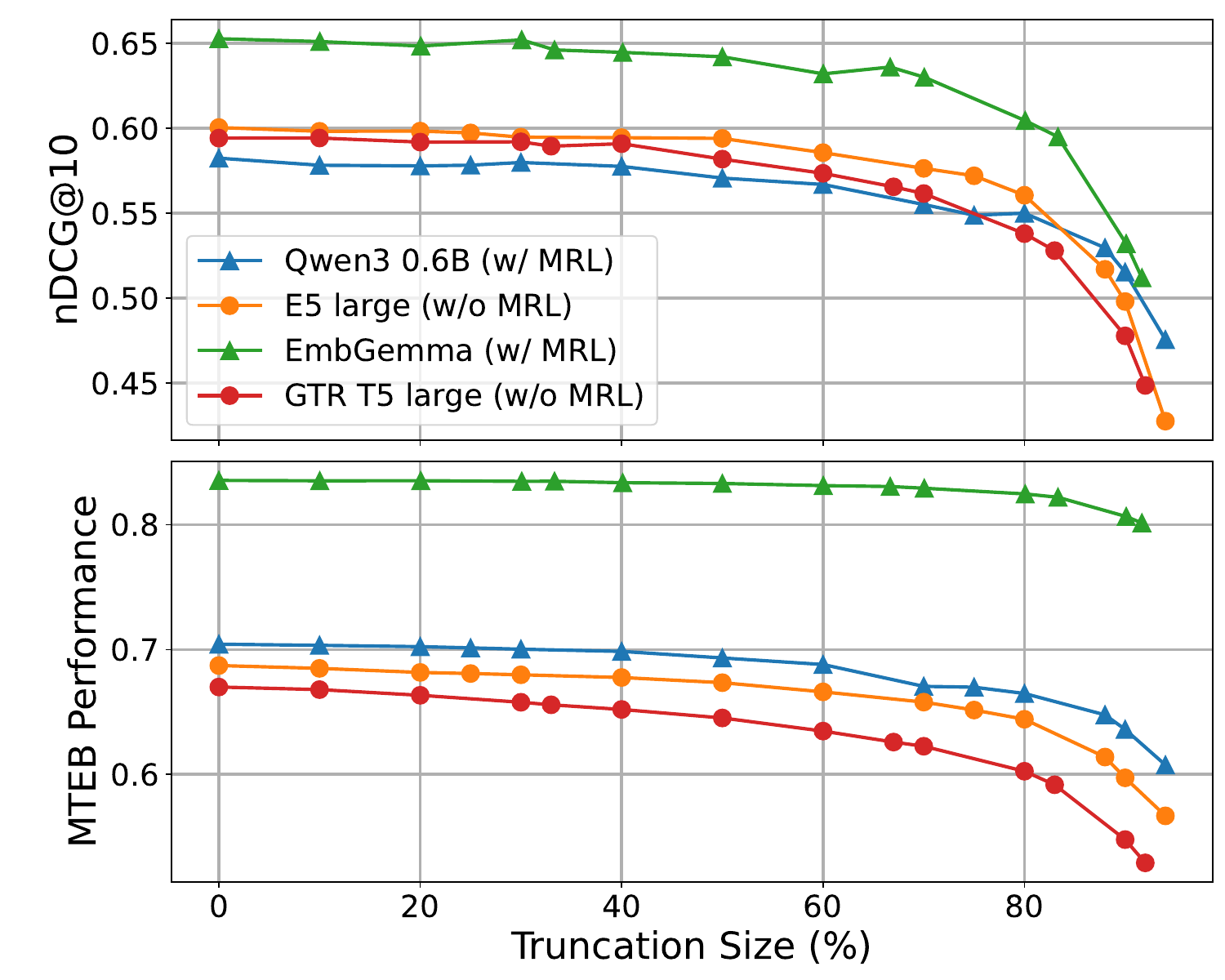}
    \caption{
        Absolute performance on NanoBEIR (top) and MTEB (bottom) of text embeddings by smaller models 
        ($\sim$500M params.) truncated at various sizes, using four publicly available text encoders with and without MRL.
        As in Fig.~\ref{fig:relative_performance_smaller_models}, the slopes indicate no significant difference in 
        robustness to truncation in MRL and non-MRL models.
        However, we do see that EmbeddingGemma scores significantly higher than all other models in both
        NanoBEIR and MTEB. 
        Such higher performance may explain EmbeddingGemma's higher robustness to truncation in MTEB, 
        as it may be easier to retain high performance after truncation when performance was high in the 
        first place.
        But this too does not generally hold, as Qwen is more robust to truncation than EmbeddingGemma in
        NanoBEIR (see Fig.~\ref{fig:relative_performance_smaller_models}), despite underperforming in absolute terms.
    }
    \label{fig:open-small-beir-mteb-absolute}
\end{figure}

\begin{figure}[h!]
    \centering
    \includegraphics[width=0.6\columnwidth]{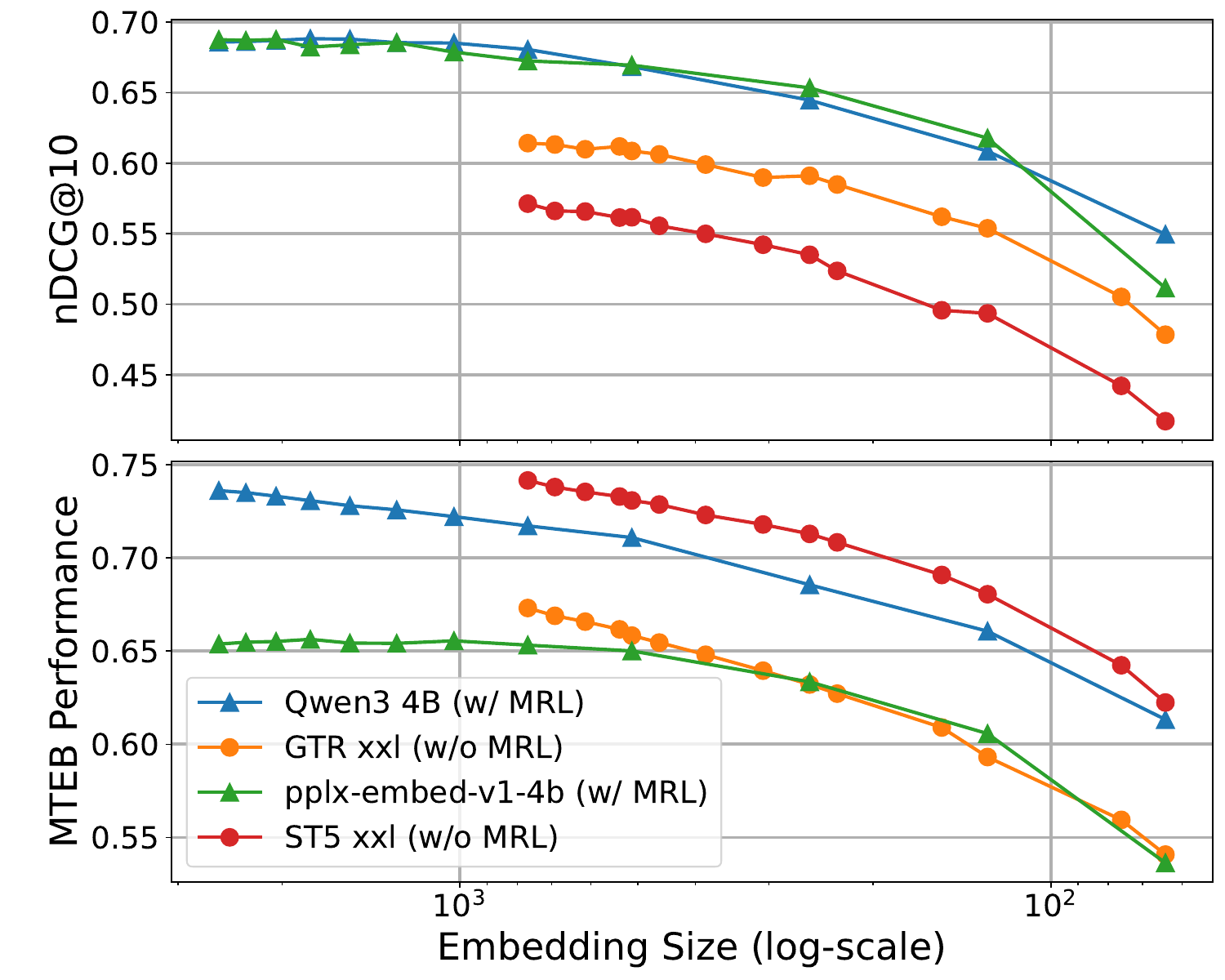}
    \caption{
        Absolute performance on NanoBEIR (top) and MTEB (bottom) of text embeddings by larger models ($\sim$4B params.)
        truncated at various sizes, using four publicly available text encoders with and without MRL.
        Similar to smaller models (see Figs.~\ref{fig:relative_performance_smaller_models} and~\ref{fig:open-small-beir-mteb-absolute}),
        the slopes indicate that MRL models are not more robust to truncation compared to non-MRL models.
    }
    \label{fig:open-large-beir-mteb-absolute}
\end{figure}

\clearpage
\section{Absolute Performance of Trained Models}\label{sec:appendix_performance_trained_models}
\begin{figure}[h!]
    \centering
    \includegraphics[width=0.95\columnwidth]{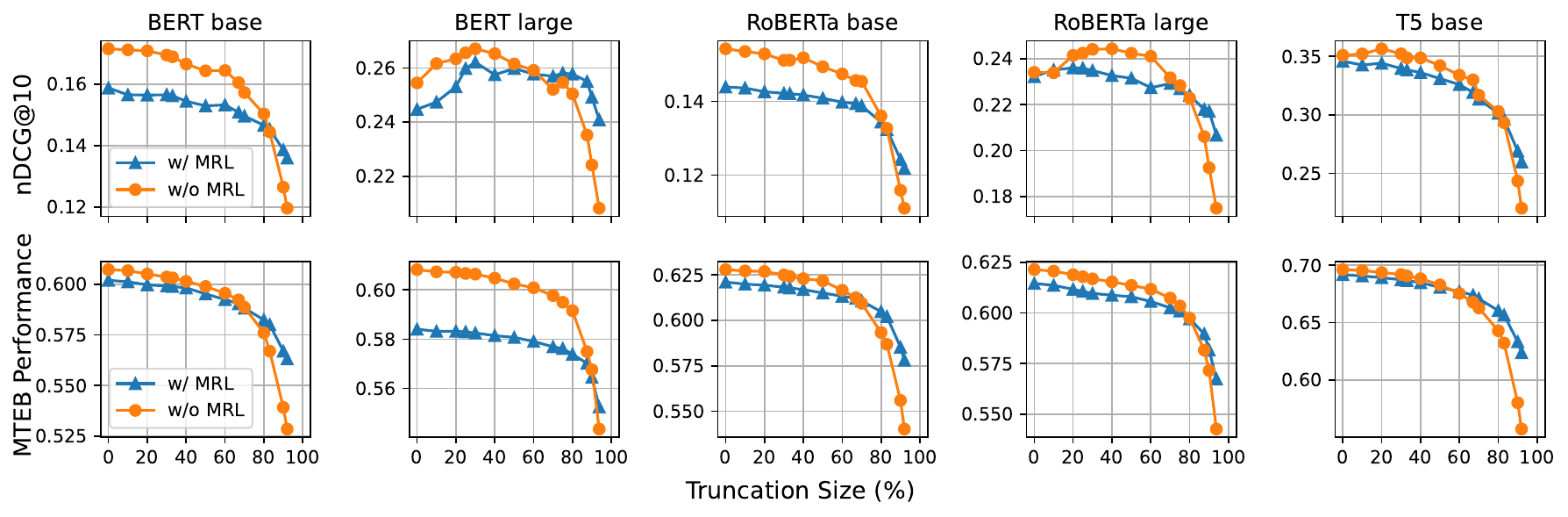}
    \caption{
        Performance on BEIR and MTEB benchmarks of five pairs of encoders trained with and without MRL.
        Non-MRL models outperform MRL models up until 80\% truncation, where MRL models become more
        beneficial.
    }
    \label{fig:trained-models-all}
\end{figure}

\section{Relative Performance of Trained Models}\label{sec:appendix_relative_performance_trained_models}
\begin{figure}[h!]
    \centering
    \includegraphics[width=0.95\columnwidth]{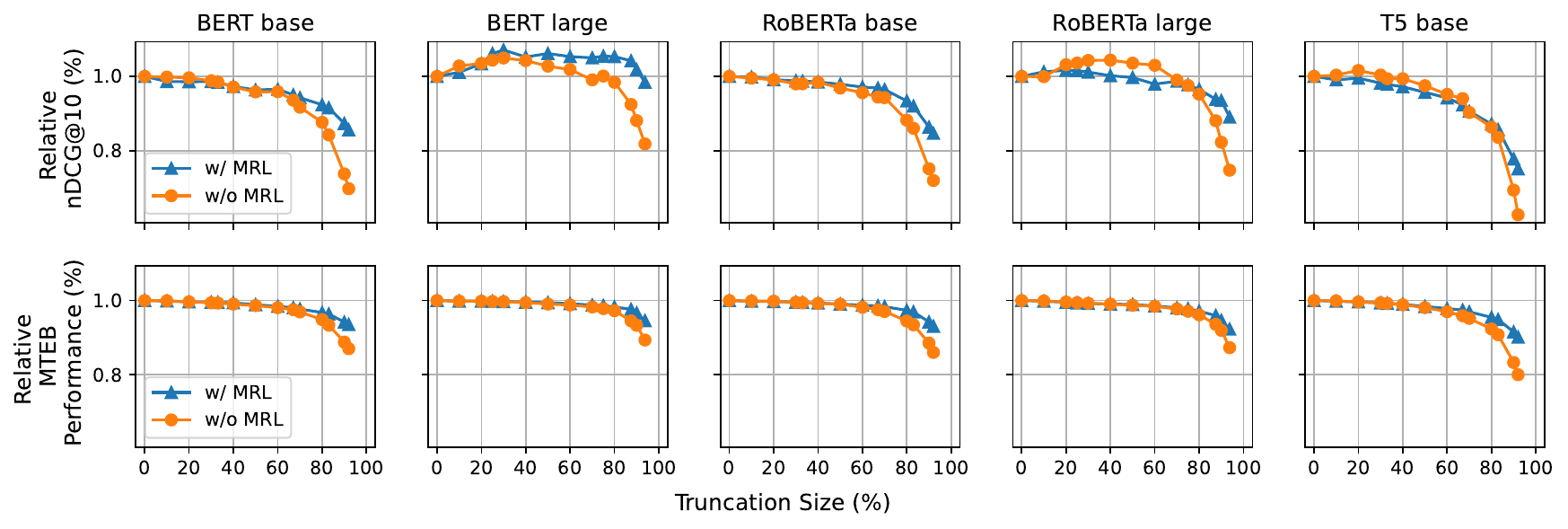}
    \caption{
        Relative performance on BEIR and MTEB benchmarks of five pairs of encoders trained with and without MRL.
    }
    \label{fig:trained-models-all-relative}
\end{figure}

\clearpage
\section{Performance of Open Models on Individual Datasets}\label{sec:appendix_performance_open_per_dataset}

% beir
\begin{figure}[h!]
    \centering
    \includegraphics[width=0.8\columnwidth]{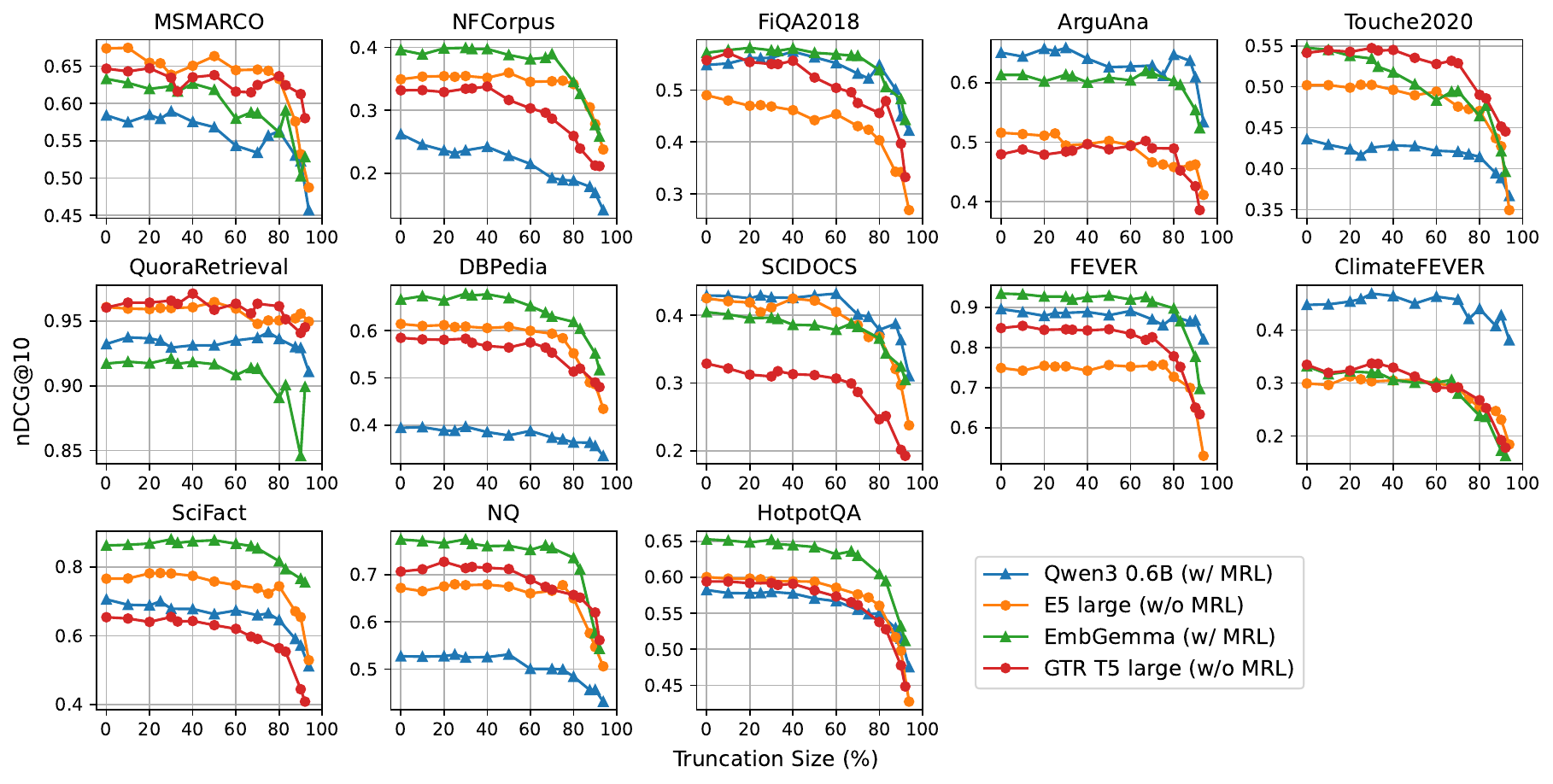}
    \caption{Performance of smaller open text encoders in NanoBEIR datasets.
    }
    \label{fig:19_beir_open-large_per-dataset_absolute}
\end{figure}
\begin{figure}[h!]
    \centering
    \includegraphics[width=0.8\columnwidth]{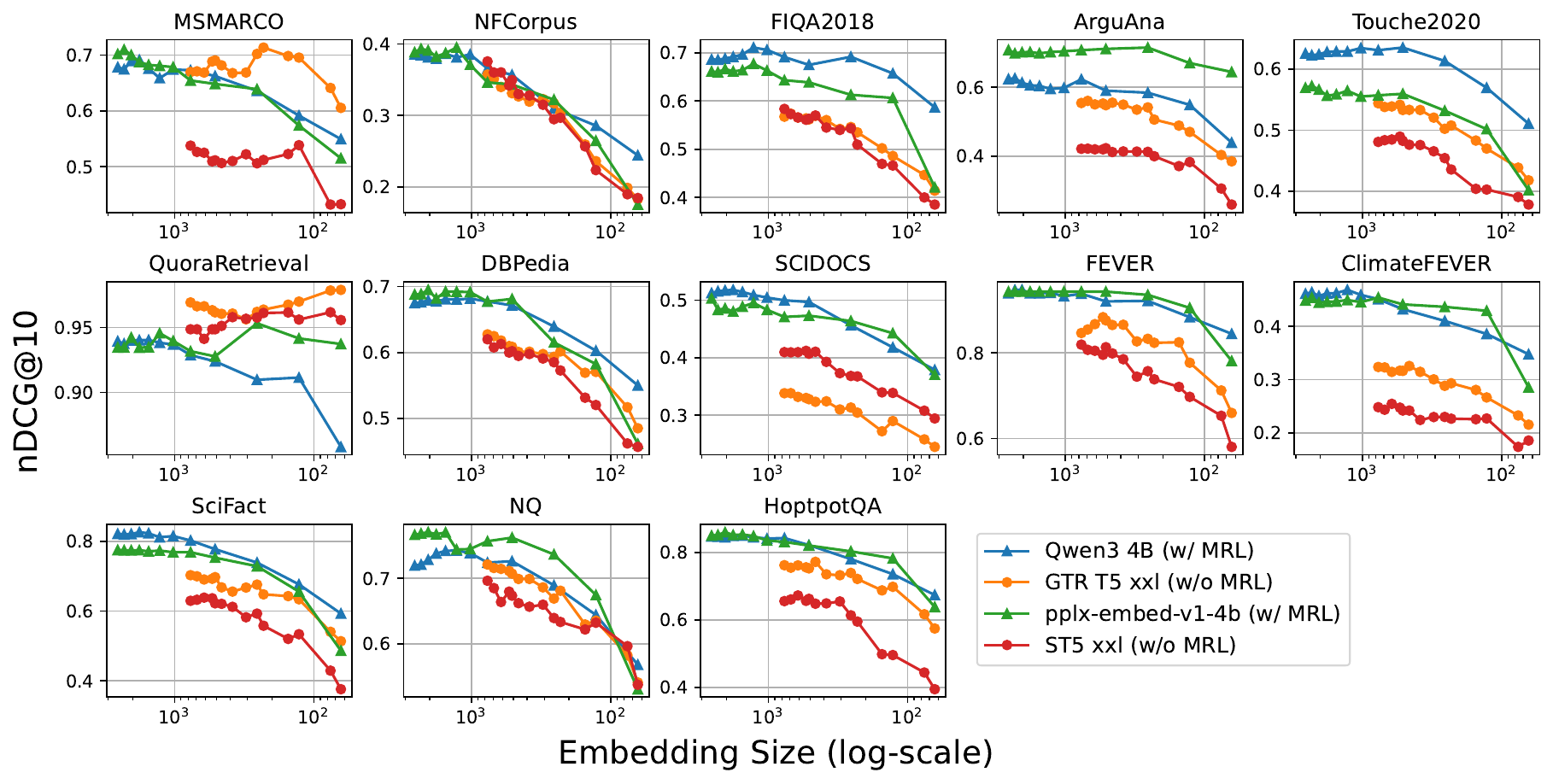}
    \caption{Performance of larger open text encoders in NanoBEIR datasets.}
    \label{fig:23_beir_open-large_per-dataset_absolute}
\end{figure}

% mteb
\begin{figure}[h!]
    \centering
    \includegraphics[width=0.8\columnwidth]{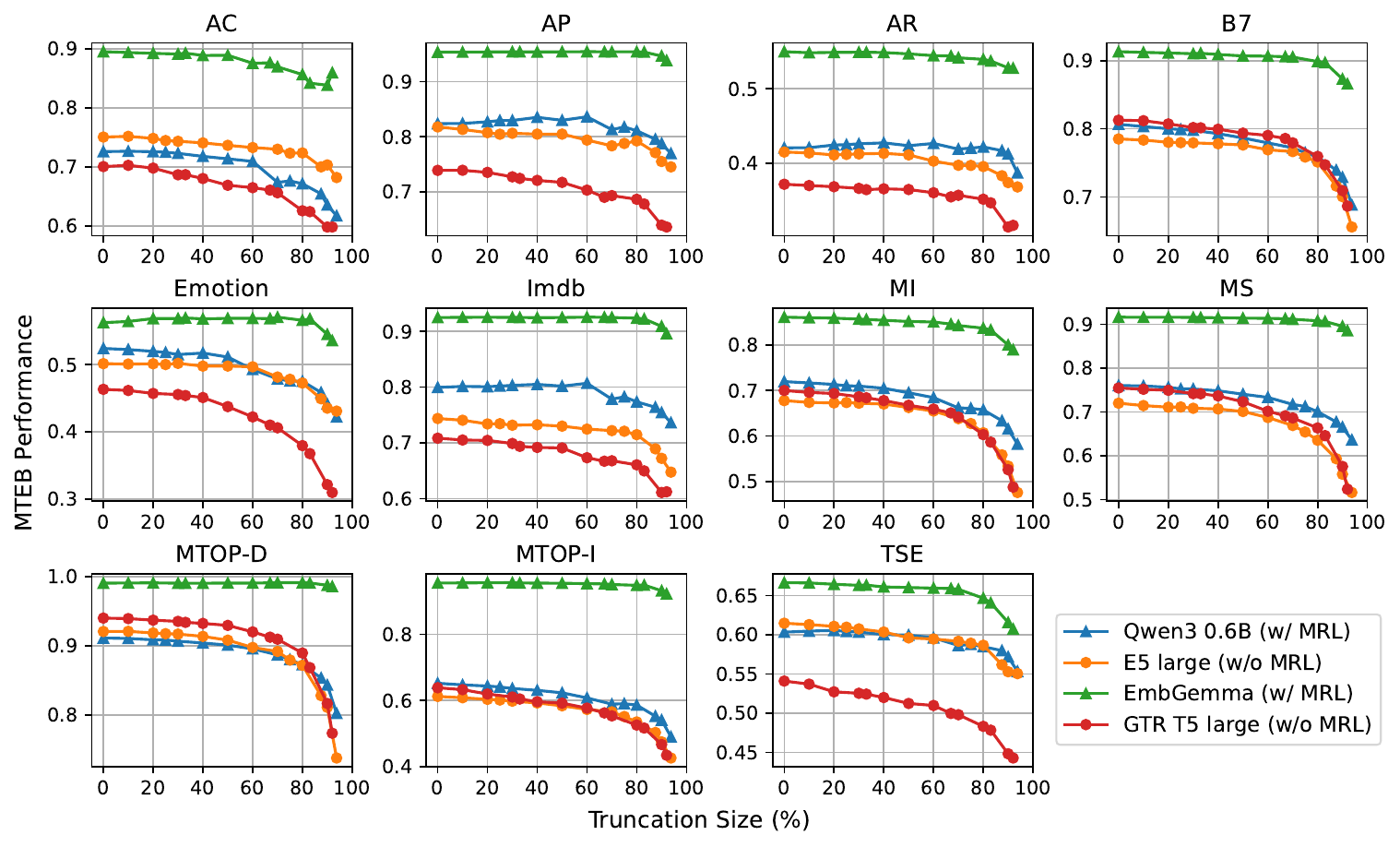}
    \caption{Performance of smaller open text encoders in MTEB datasets.}
    \label{fig:07_mteb_open-small_per-dataset_absolute}
\end{figure}
\begin{figure}[h!]
    \centering
    \includegraphics[width=0.8\columnwidth]{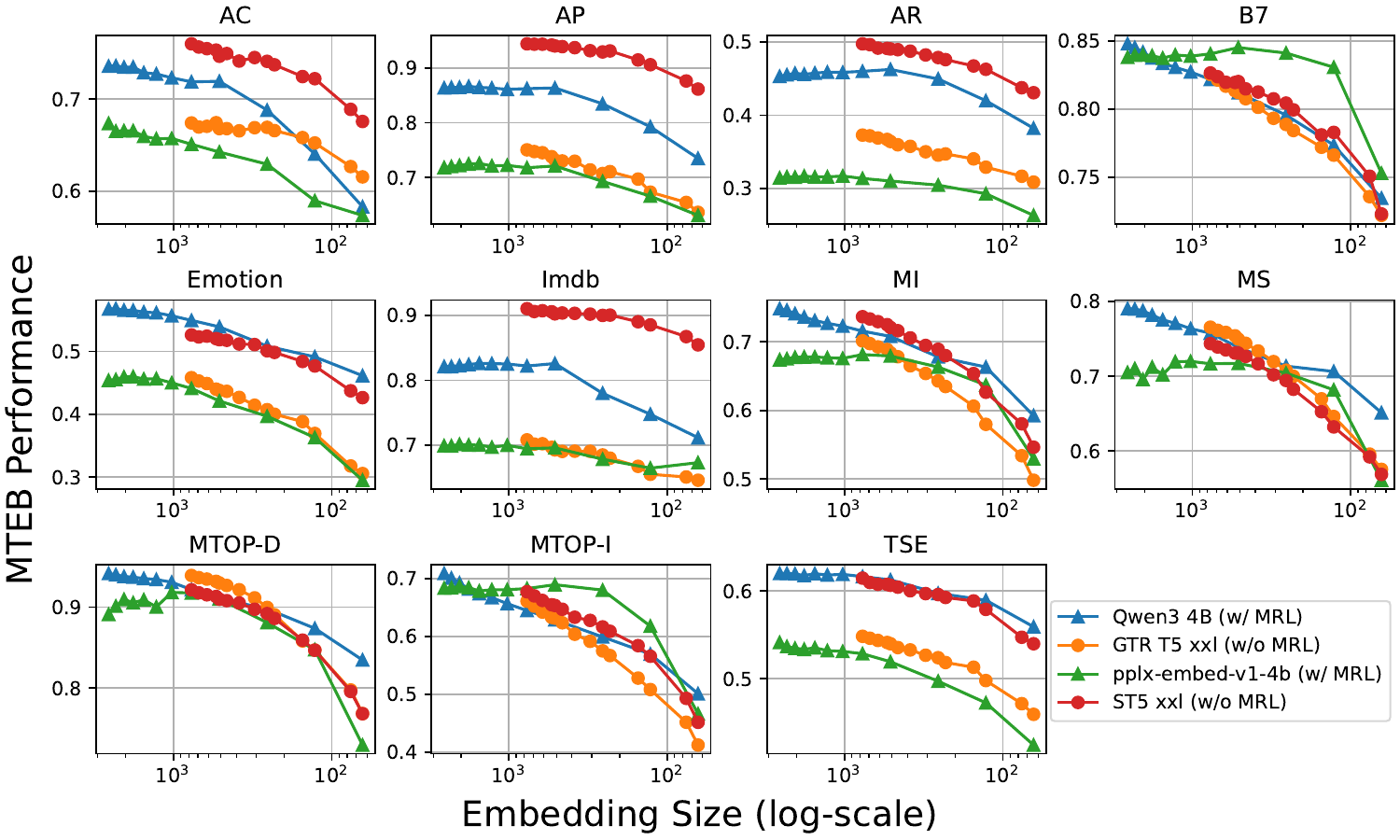}
    \caption{Performance of larger open text encoders in MTEB datasets.}
    \label{fig:11_mteb_open-large_dataset-per_absolute}
\end{figure}

\clearpage
\section{Performance of Trained Models on Individual Datasets}\label{sec:appendix_performance_trained_per_dataset}

% beir
\begin{figure}[h!]
    \centering
    \includegraphics[width=0.8\columnwidth]{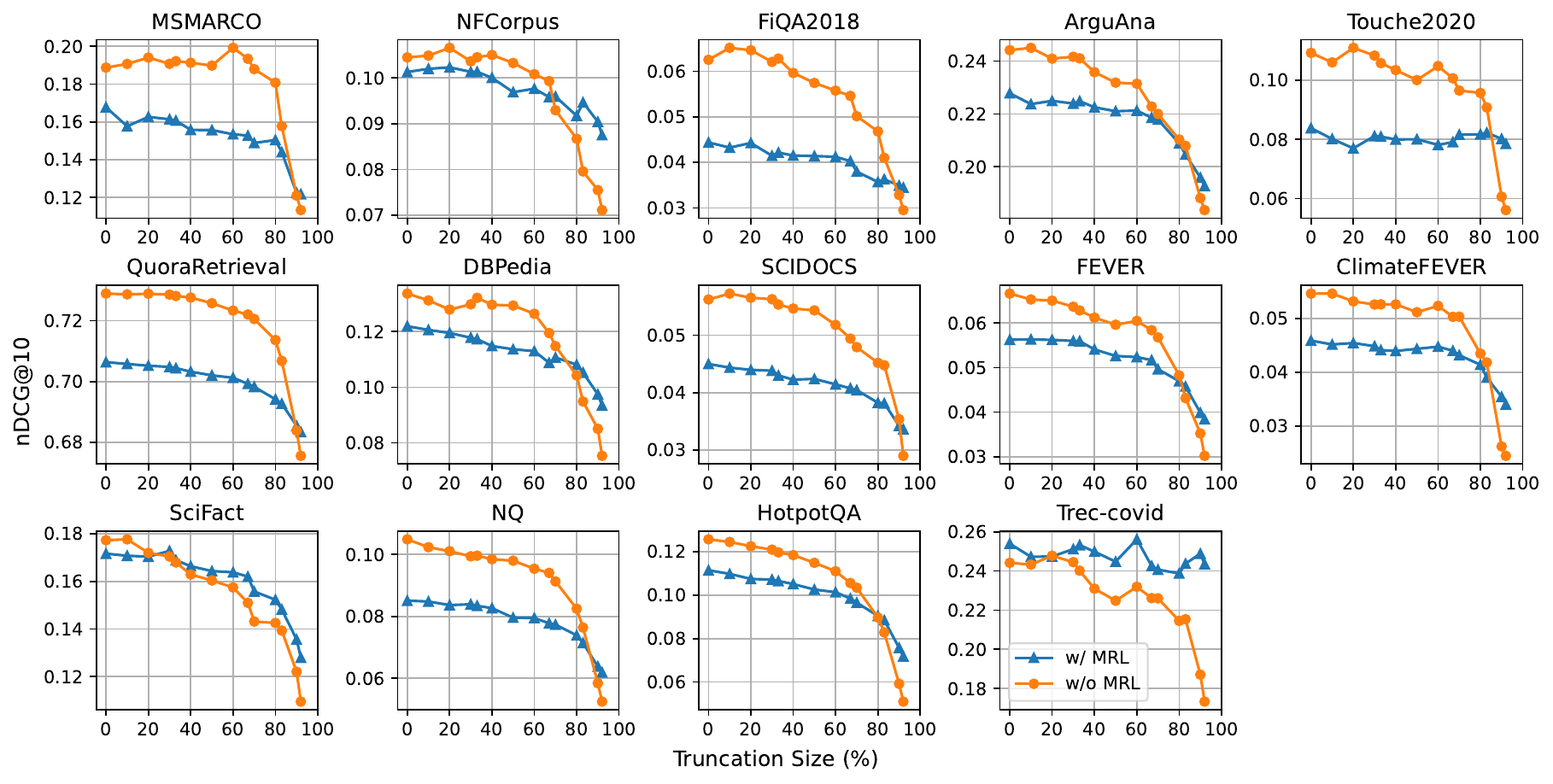}
    \caption{BERT base performance on each of the BEIR datasets. 
    }
    \label{fig:15-1_beir_ours-bert-base_per-dataset_absolute}
\end{figure}
\begin{figure}[h!]
    \centering
    \includegraphics[width=0.8\columnwidth]{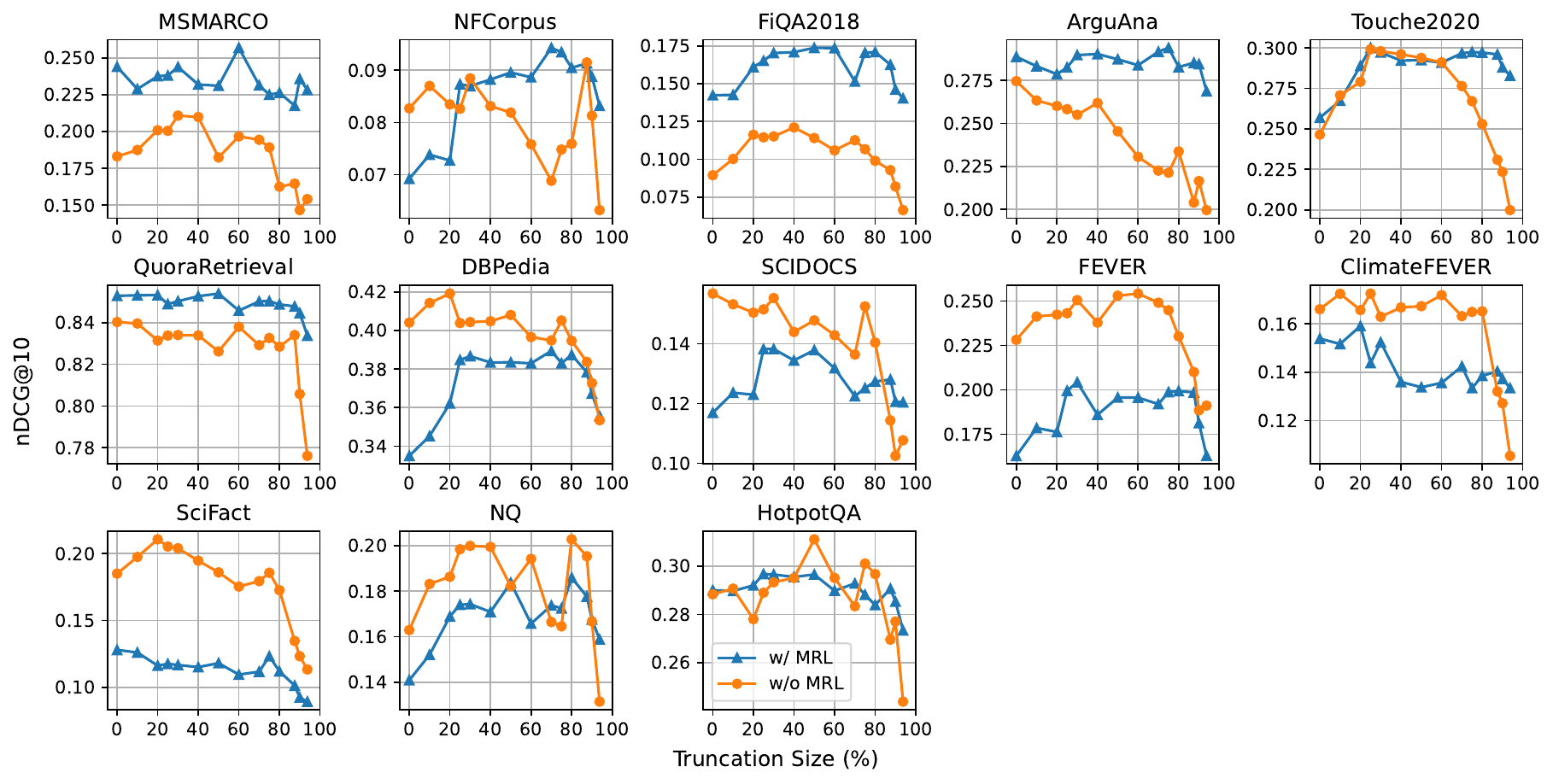}
    \caption{BERT large performance on each of the NanoBEIR datasets.}
    \label{fig:15-2_beir_ours-bert-large_per-dataset_absolute}
\end{figure}
\begin{figure}[h!]
    \centering
    \includegraphics[width=0.8\columnwidth]{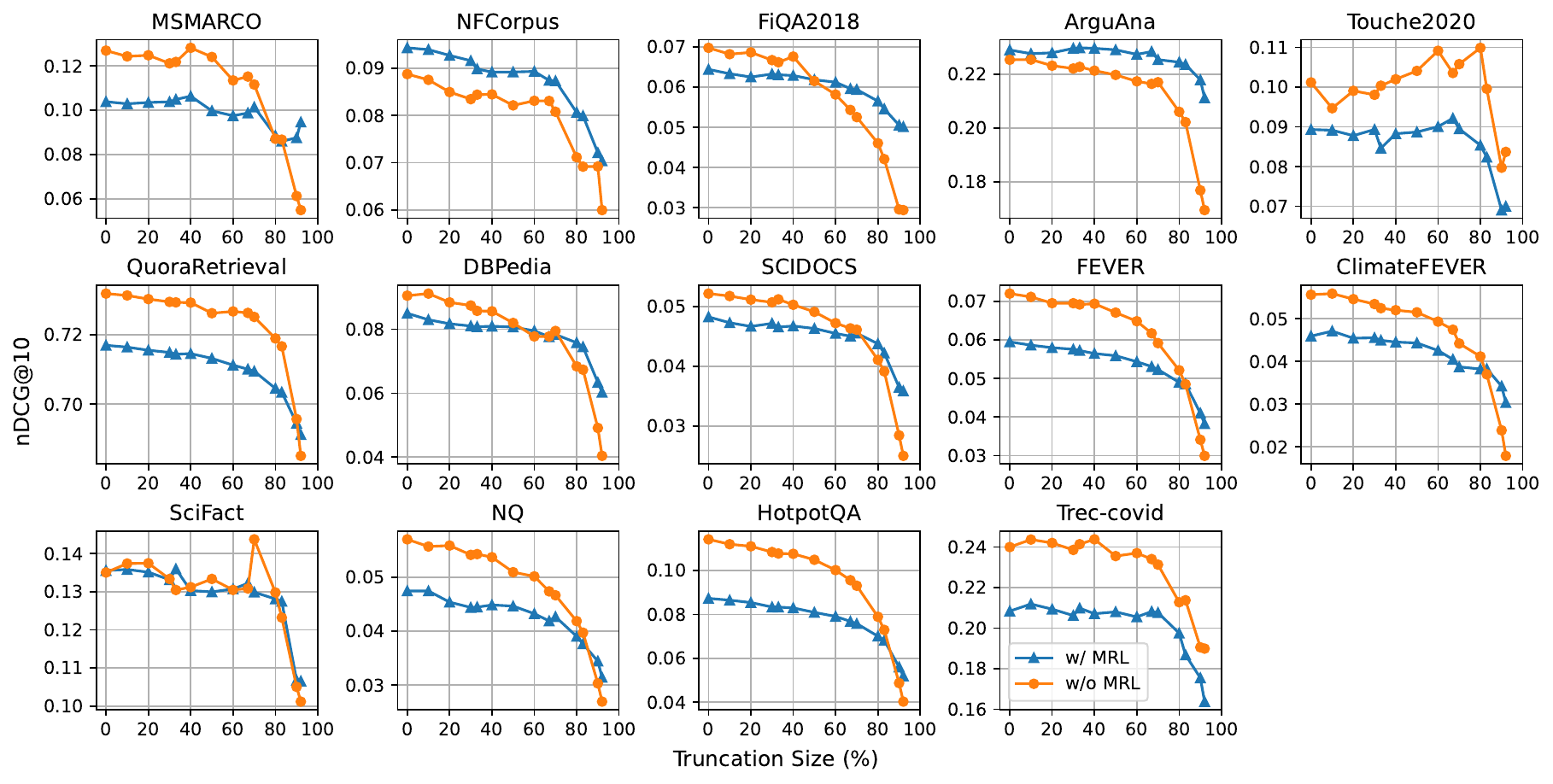}
    \caption{RoBERTa base performance on each of the BEIR datasets.}
    \label{fig:15-3_beir_ours-roberta-base_per-dataset_absolute}
\end{figure}
\begin{figure}[h!]
    \centering
    \includegraphics[width=0.8\columnwidth]{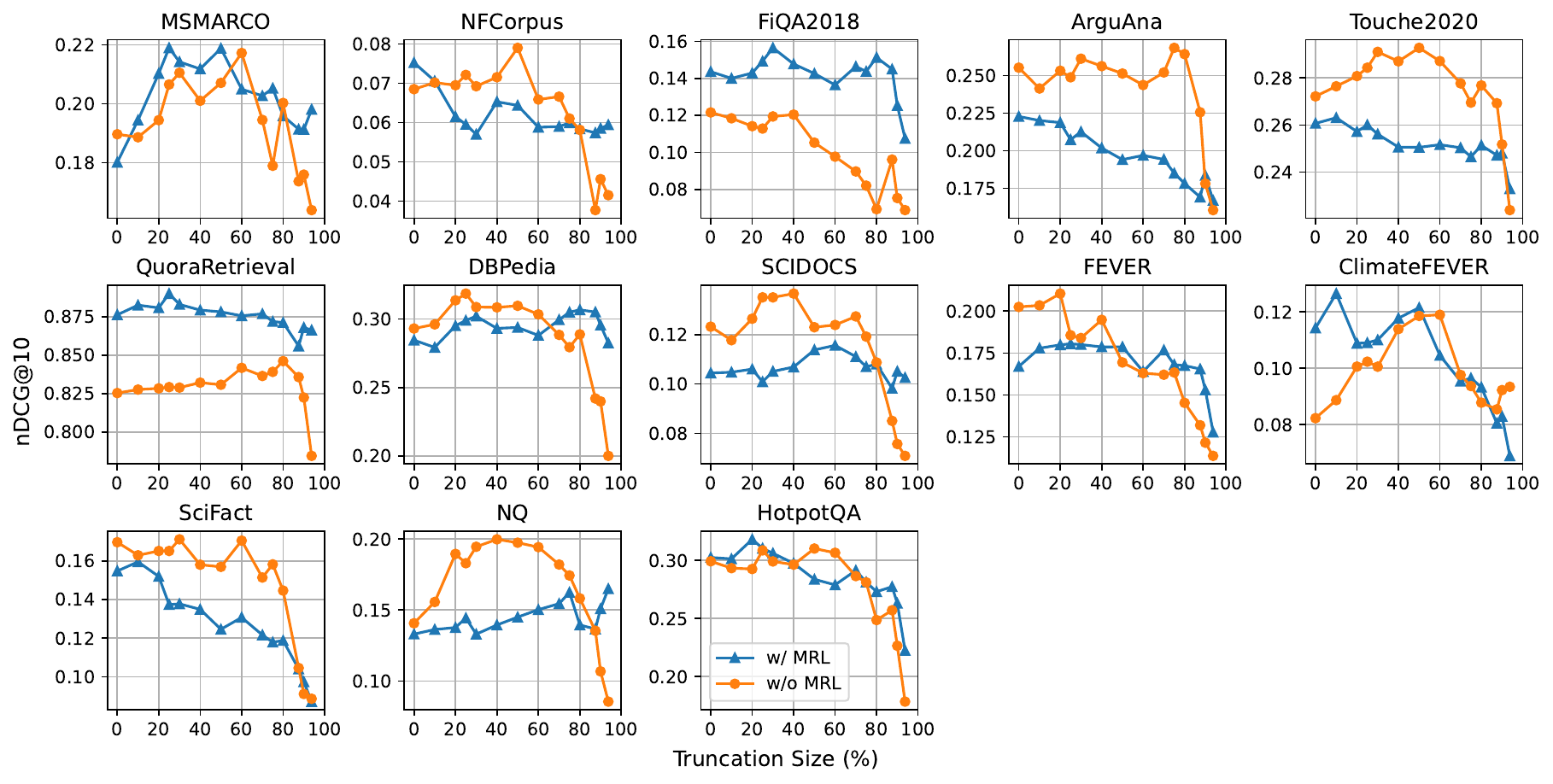}
    \caption{RoBERTa large performance on each of the NanoBEIR datasets.}
    \label{fig:15-4_beir_ours-roberta-large_per-dataset_absolute}
\end{figure}
\begin{figure}[h!]
    \centering
    \includegraphics[width=0.8\columnwidth]{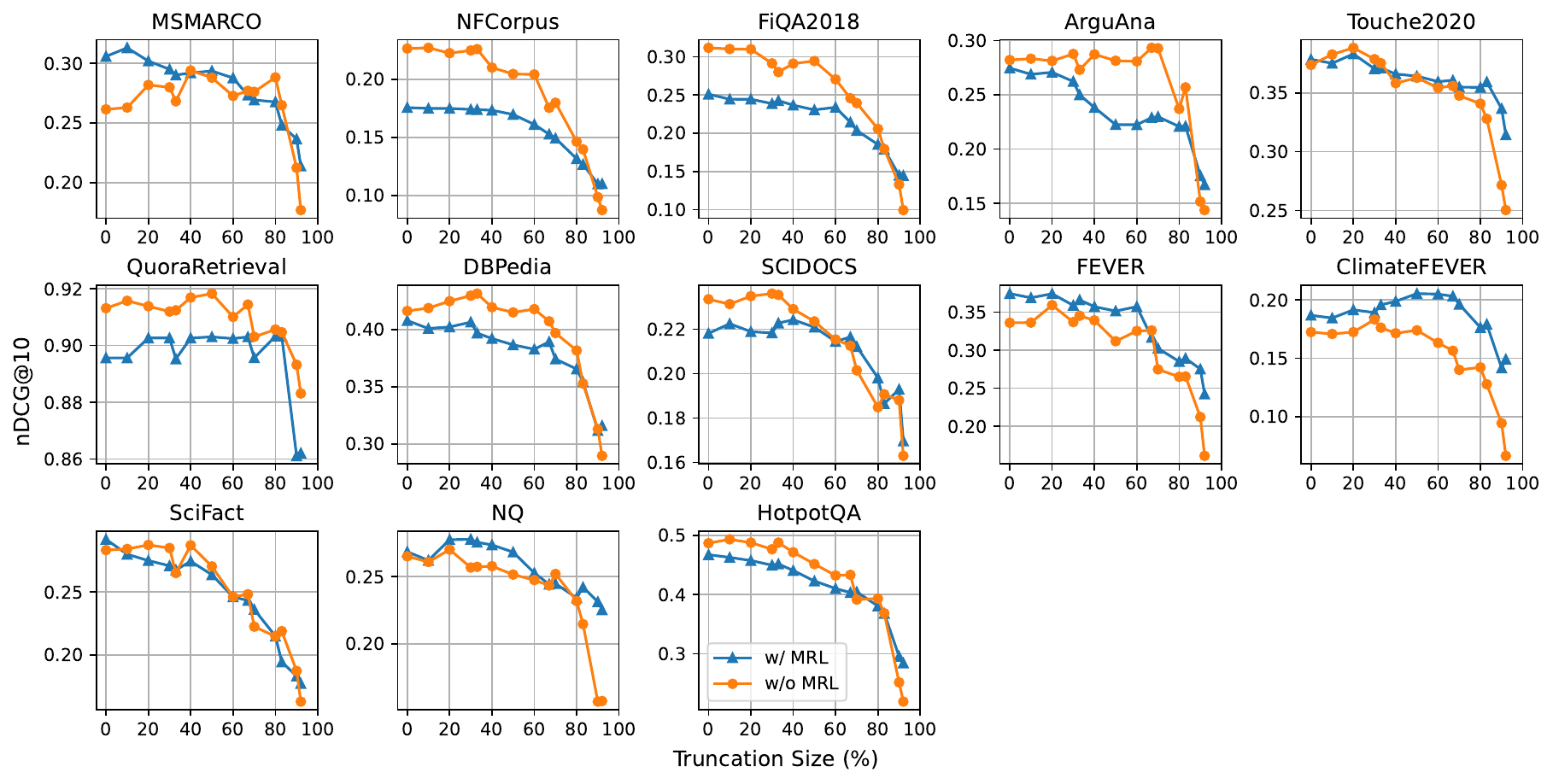}
    \caption{T5 base performance on each of the NanoBEIR datasets.}
    \label{fig:15-5_beir_ours-t5-base_per-dataset_absolute}
\end{figure}

% mteb
\begin{figure}[h!]
    \centering
    \includegraphics[width=0.8\columnwidth]{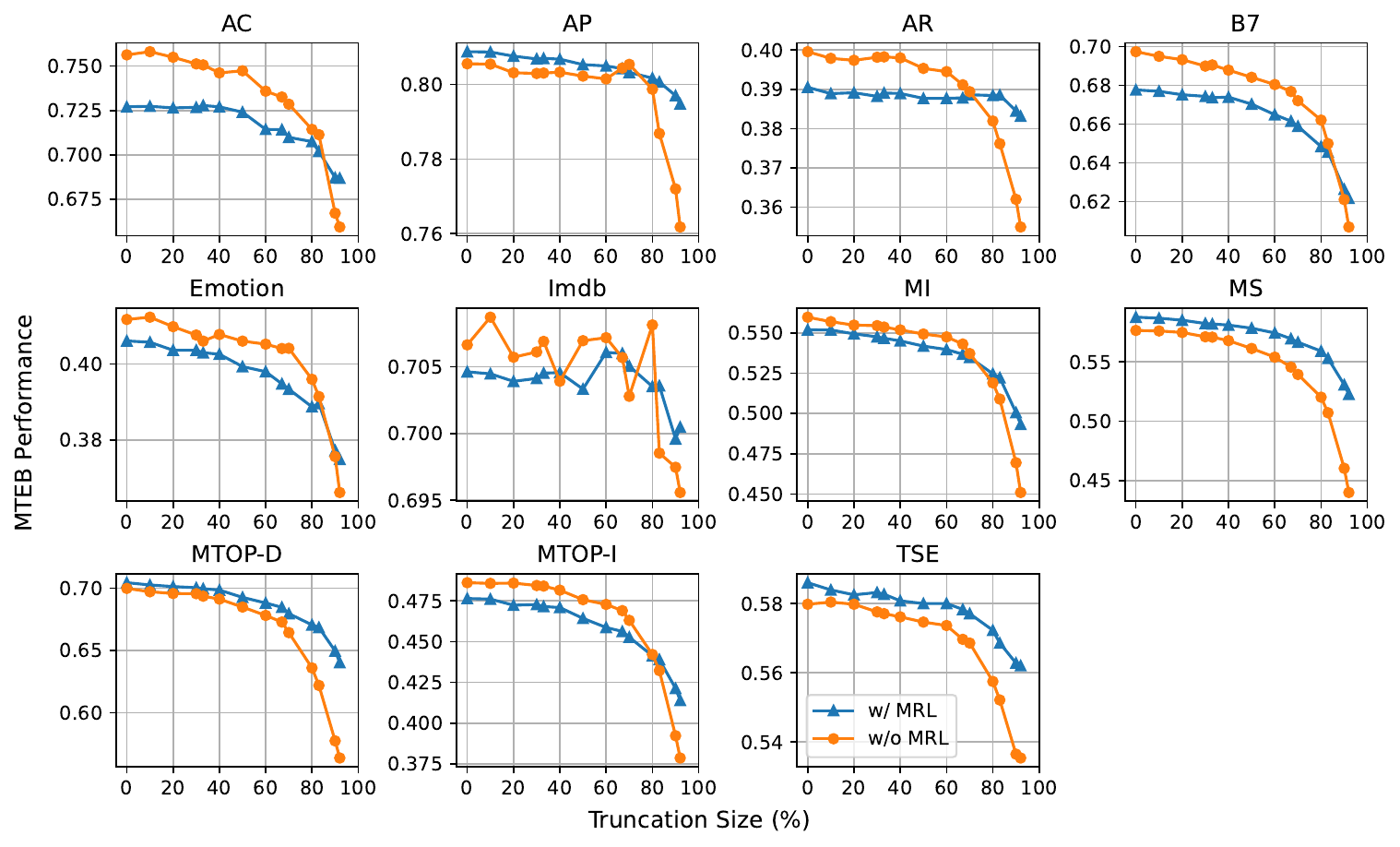}
    \caption{BERT base performance on each of the MTEB datasets.}
    \label{fig:03-1_mteb_ours-bert-base_per-dataset_absolute}
\end{figure}
\begin{figure}[h!]
    \centering
    \includegraphics[width=0.8\columnwidth]{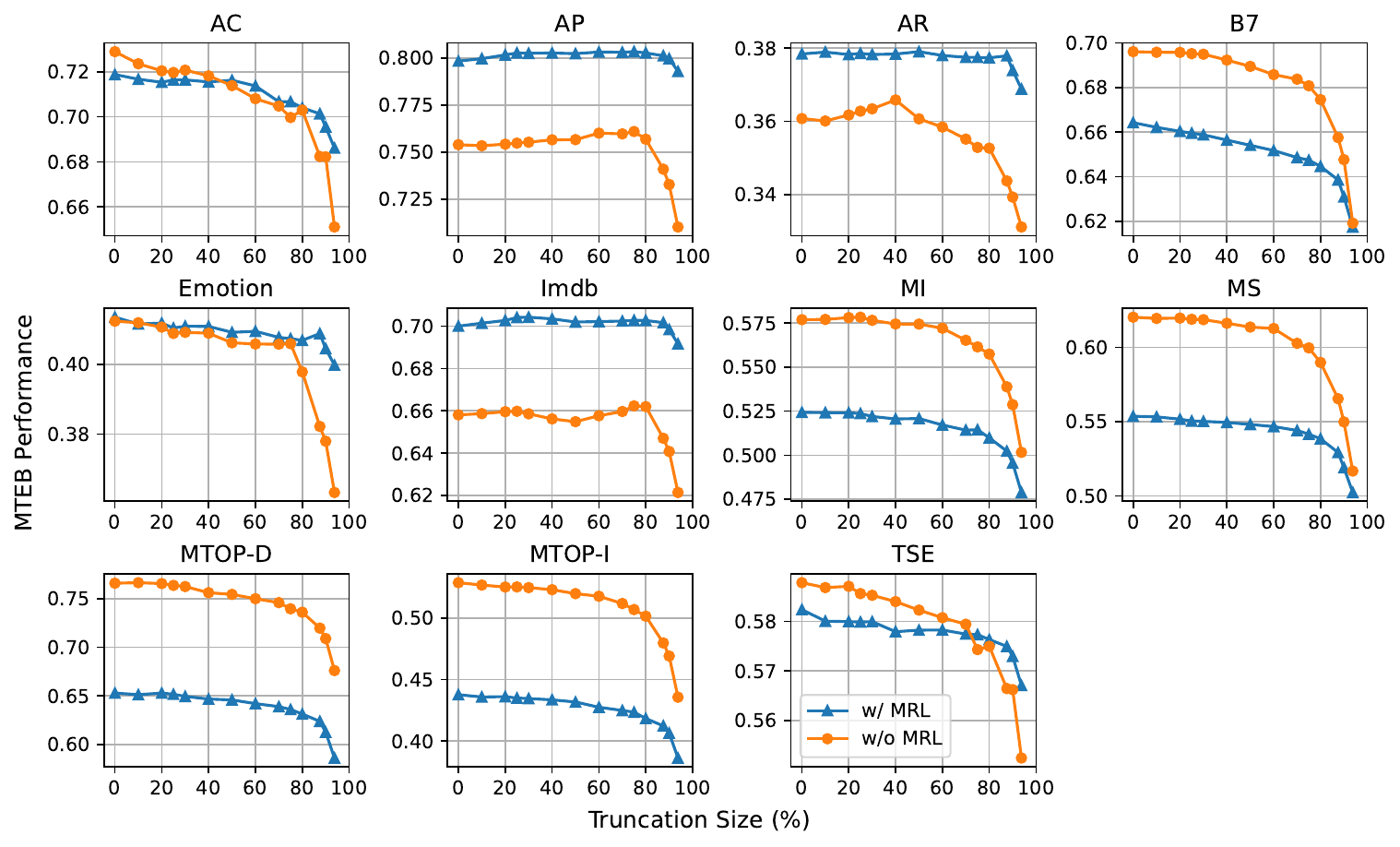}
    \caption{BERT large performance on each of the MTEB datasets.}
    \label{fig:03-2_mteb_ours-bert-large_per-dataset_absolute}
\end{figure}
\begin{figure}[h!]
    \centering
    \includegraphics[width=0.8\columnwidth]{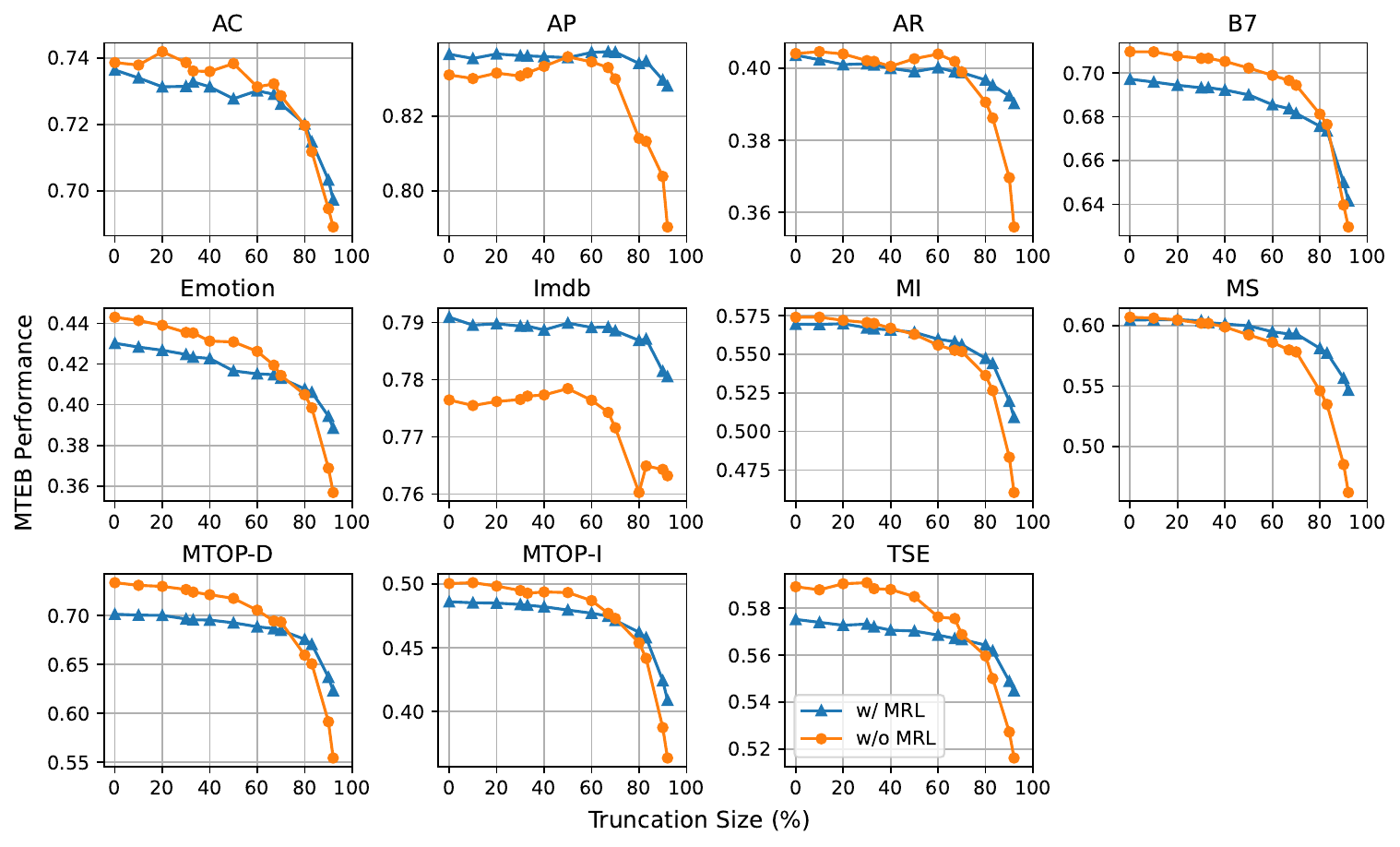}
    \caption{RoBERTa base performance on each of the MTEB datasets.}
    \label{fig:03-3_mteb_ours-roberta-base_per-dataset_absolute}
\end{figure}
\begin{figure}[h!]
    \centering
    \includegraphics[width=0.8\columnwidth]{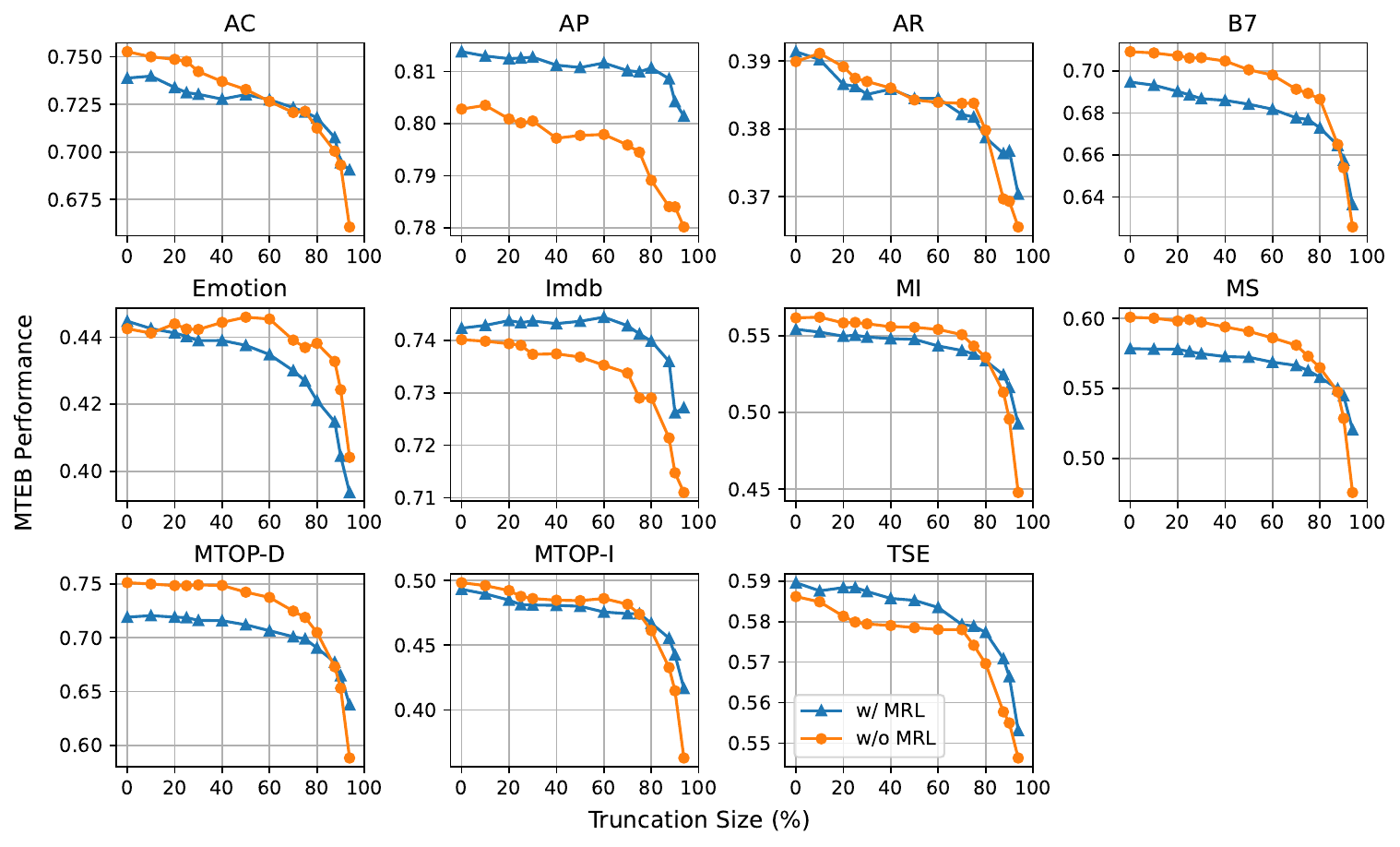}
    \caption{RoBERTa large performance on each of the MTEB datasets.}
    \label{fig:03-4_mteb_ours-roberta-large_per-dataset_absolute}
\end{figure}
\begin{figure}[h!]
    \centering
    \includegraphics[width=0.8\columnwidth]{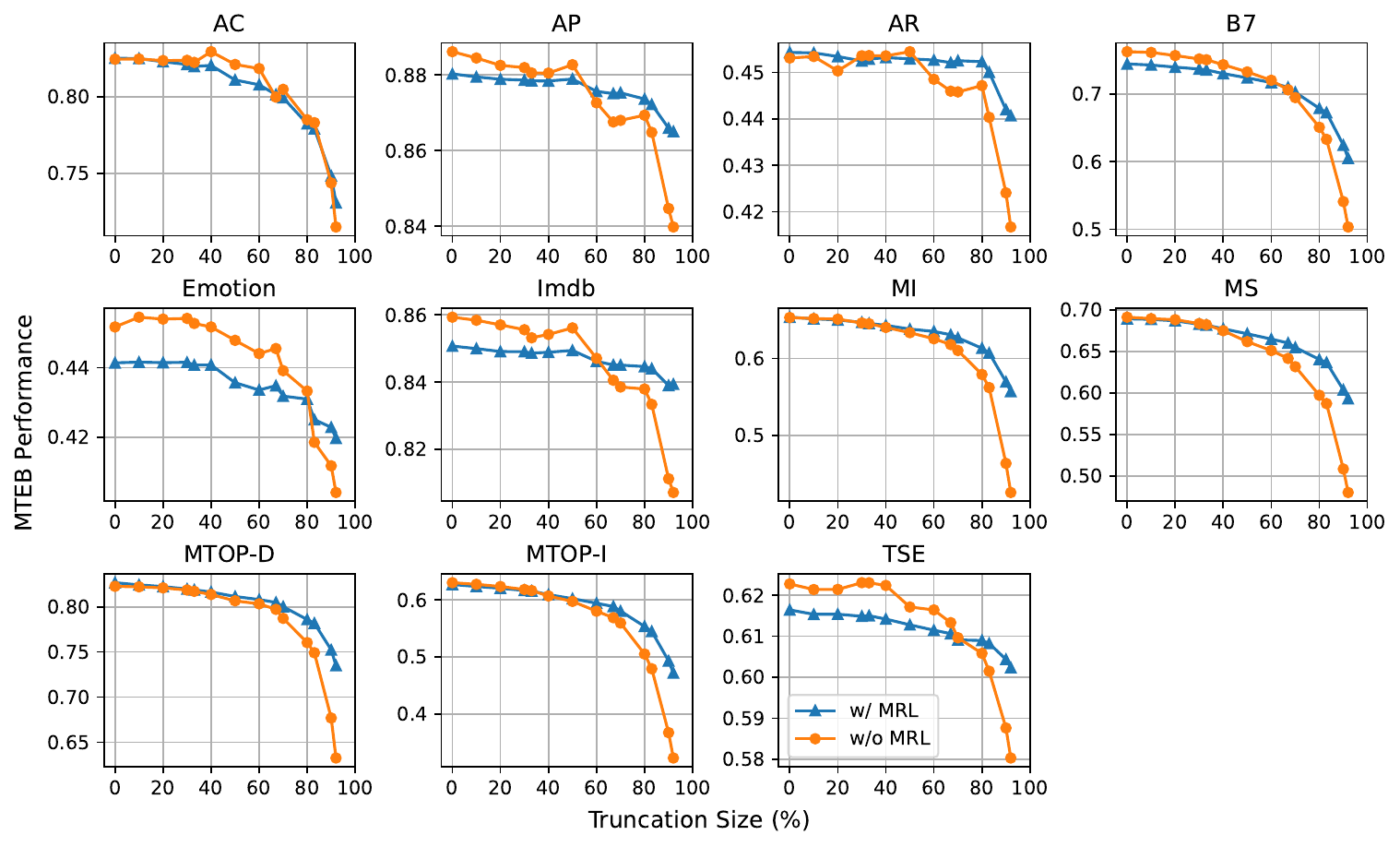}
    \caption{T5 base performance on each of the MTEB datasets.}
    \label{fig:03-5_mteb_ours-t5-base_per-dataset_absolute}
\end{figure}

\clearpage
\section{Standard Deviations of Each Dimension}\label{sec:appendix_std_per_dimension}
\begin{figure}[h!]
    \centering
    \includegraphics[width=1.0\columnwidth]{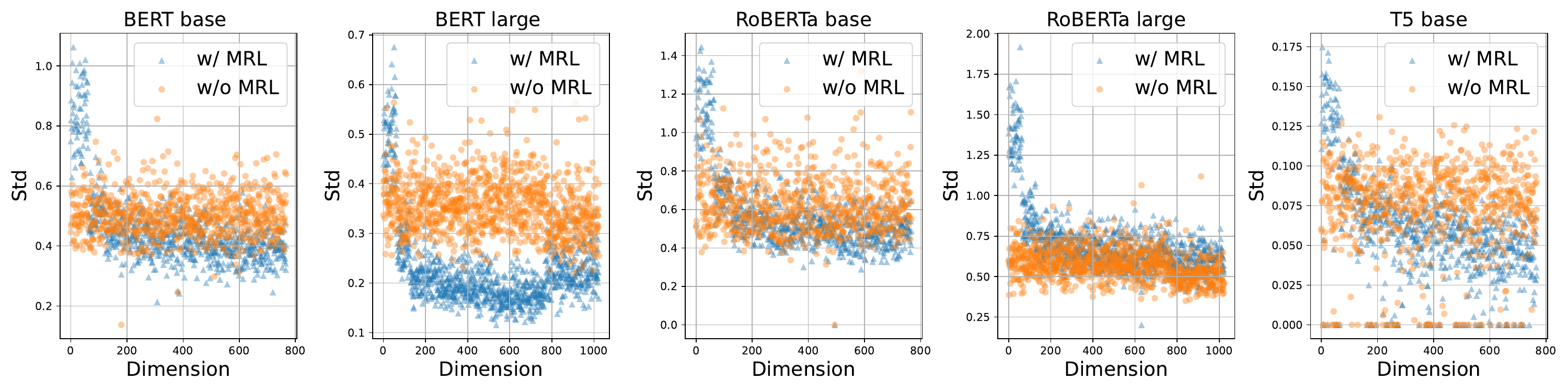}
    \caption{
        Standard deviations of values taken by each dimension when encoding different texts.
        We observe that MRL models seem to store more information in the lower dimensions 
        compared to non-MRL models, which may be related to its better performance at heavy
        truncation levels.
    }
    \label{fig:std-all-models}
\end{figure}

\section{Computation of crossover point}
\label{sec:crossover}
We use linear interpolation to estimate the crossover point between each 
MRL and non-MRL model pair based on the results in 
Sec.~\ref{sec:trained-models} (summarized in Fig.~\ref{fig:trained-models-all}).
Through this analysis, we observe that the crossovers occur on average
at approximately 75\% truncation across the two benchmarks and
five model pairs.
Since all experiments use a truncation step size of 10\%,
we conservatively summarize this finding by stating that
non-MRL models remain comparable to MRL models until roughly
80\% of the embedding dimensions are truncated.

\clearpage
\section{Confidence Intervals}

For precision, and following~\citet{senthil2026cohyde}, we report in Tables~\ref{tab:ci-base} 
and~\ref{tab:ci-large}, for each truncation level, the per-query 95\% confidence interval (CI) 
for NDCG@10, computed via percentile bootstrap using 10000 samples with replacement over a 
fixed seed. 
In addition, to determine the statistical significance of the observed performance difference
at each truncation level, we tested whether the 95\% CIs of the per-query difference excludes 
zero, the null hypothesis of no difference ($p < 0.05$). Each underlined entry in 
Tables~\ref{tab:ci-base} and~\ref{tab:ci-large} shows when the difference is statistically 
significant. 
For RoBERTa and T5, the difference is only statistically significant at heavy truncation, 
such as 64 remaining dimensions for RoBERTa-base (8\% of the original embedding size).
The exception is BERT, where no MRL seems more favourable most of the time.

\begin{table}[h!]
    \centering
    \small
    \setlength{\tabcolsep}{1.2pt}
    \begin{tabular}{ccllllllllllllll}
    \toprule
     &  & \multicolumn{1}{c}{\textbf{64}} & \multicolumn{1}{c}{\textbf{76}} & \multicolumn{1}{c}{\textbf{128}} & \multicolumn{1}{c}{\textbf{153}} & \multicolumn{1}{c}{\textbf{256}} & \multicolumn{1}{c}{\textbf{307}} & \multicolumn{1}{c}{\textbf{384}} & \multicolumn{1}{c}{\textbf{460}} & \multicolumn{1}{c}{\textbf{512}} & \multicolumn{1}{c}{\textbf{537}} & \multicolumn{1}{c}{\textbf{614}} & \multicolumn{1}{c}{\textbf{691}} & \multicolumn{1}{c}{\textbf{768}} \\
    \midrule
    \multirow{2}{*}{\rotatebox[origin=c]{90}{\textbf{B-b}}} & \textbf{w/} & $0.22^{0.25}_{0.20}$ & $0.23^{0.25}_{0.20}$ & $0.24^{0.26}_{0.21}$ & $0.24^{0.26}_{0.21}$ & $0.25^{0.27}_{0.22}$ & $0.25^{0.28}_{0.23}$ & $0.26^{0.28}_{0.23}$ & $0.26^{0.28}_{0.23}$ & $0.26^{0.28}_{0.23}$ & $0.26^{0.28}_{0.23}$ & $0.26^{0.28}_{0.23}$ & $0.26^{0.28}_{0.23}$ & $0.26^{0.28}_{0.23}$ \\
    [2pt]
     & \textbf{w/o} & $0.22^{0.25}_{0.20}$ & $0.23^{0.25}_{0.20}$ & $0.25^{0.28}_{0.23}$ & $0.26^{0.28}_{0.23}$ & \underline{$0.27$}$^{0.30}_{0.25}$ & \underline{$0.28$}$^{0.31}_{0.25}$ & \underline{$0.28$}$^{0.30}_{0.25}$ & \underline{$0.28$}$^{0.31}_{0.25}$ & \underline{$0.28$}$^{0.31}_{0.25}$ & \underline{$0.28$}$^{0.31}_{0.25}$ & \underline{$0.28$}$^{0.31}_{0.26}$ & \underline{$0.28$}$^{0.31}_{0.26}$ & \underline{$0.28$}$^{0.31}_{0.26}$ \\
    \midrule

    \multirow{2}{*}{\rotatebox[origin=c]{90}{\textbf{R-b}}} & \textbf{w/} & \underline{$0.23$}$^{0.25}_{0.20}$ & $0.23^{0.25}_{0.21}$ & $0.24^{0.27}_{0.22}$ & $0.24^{0.27}_{0.22}$ & $0.25^{0.27}_{0.22}$ & $0.25^{0.28}_{0.23}$ & $0.25^{0.28}_{0.23}$ & $0.26^{0.28}_{0.23}$ & $0.26^{0.28}_{0.23}$ & $0.26^{0.28}_{0.23}$ & $0.26^{0.29}_{0.23}$ & $0.26^{0.28}_{0.23}$ & $0.26^{0.28}_{0.23}$ \\
    [2pt]
     & \textbf{w/o} & $0.21^{0.23}_{0.18}$ & $0.22^{0.24}_{0.19}$ & $0.24^{0.26}_{0.21}$ & $0.24^{0.27}_{0.22}$ & $0.26^{0.29}_{0.24}$ & $0.26^{0.29}_{0.24}$ & $0.26^{0.29}_{0.24}$ & $0.27^{0.29}_{0.24}$ & $0.27^{0.29}_{0.24}$ & $0.27^{0.30}_{0.24}$ & $0.27^{0.30}_{0.24}$ & $0.27^{0.30}_{0.25}$ & $0.27^{0.30}_{0.25}$ \\

    \midrule

    \multirow{2}{*}{\rotatebox[origin=c]{90}{\textbf{T-b}}} & \textbf{w/} & \underline{$0.26$}$^{0.29}_{0.24}$ & \underline{$0.27$}$^{0.30}_{0.24}$ & $0.30^{0.32}_{0.27}$ & $0.30^{0.33}_{0.28}$ & $0.32^{0.35}_{0.29}$ & $0.33^{0.35}_{0.30}$ & $0.33^{0.36}_{0.30}$ & $0.34^{0.36}_{0.31}$ & $0.34^{0.37}_{0.31}$ & $0.34^{0.37}_{0.31}$ & $0.34^{0.37}_{0.32}$ & $0.34^{0.37}_{0.32}$ & $0.35^{0.37}_{0.32}$ \\
    [2pt]
     & \textbf{w/o} & $0.22^{0.25}_{0.20}$ & $0.24^{0.27}_{0.22}$ & $0.29^{0.32}_{0.27}$ & $0.30^{0.33}_{0.28}$ & $0.33^{0.36}_{0.30}$ & $0.33^{0.36}_{0.31}$ & $0.34^{0.37}_{0.32}$ & $0.35^{0.38}_{0.32}$ & $0.35^{0.38}_{0.32}$ & $0.35^{0.38}_{0.33}$ & $0.36^{0.38}_{0.33}$ & $0.35^{0.38}_{0.33}$ & $0.35^{0.38}_{0.32}$ \\

    \bottomrule
    \end{tabular}
    \caption{95\% Confidence intervals (CIs) via per-query bootstrap for three base-sized model pairs: BERT-base (B-b), RoBERTa-base (R-b), T5-base (T-b) on NanoBEIR datasets. Each column shows the average performance and CIs for the indicated number of remaining dimensions after truncation. A score is \underline{underlined} when it significantly outperforms its counterpart by two-sided paired t-test ($p < 0.05$).}
    \label{tab:ci-base}
\end{table}
\begin{table}[h!]
    \centering
    \small
    \setlength{\tabcolsep}{2.5pt}
    \begin{tabular}{cclllllllllllll}
    \toprule
     &  & \multicolumn{1}{c}{\textbf{64}} & \multicolumn{1}{c}{\textbf{102}} & \multicolumn{1}{c}{\textbf{128}} & \multicolumn{1}{c}{\textbf{204}} & \multicolumn{1}{c}{\textbf{256}} & \multicolumn{1}{c}{\textbf{409}} & \multicolumn{1}{c}{\textbf{512}} & \multicolumn{1}{c}{\textbf{614}} & \multicolumn{1}{c}{\textbf{716}} & \multicolumn{1}{c}{\textbf{819}} & \multicolumn{1}{c}{\textbf{921}} & \multicolumn{1}{c}{\textbf{1024}} \\

    \midrule
    \multirow{2}{*}{\rotatebox[origin=c]{90}{\textbf{B-l}}} & \textbf{w/} & $0.22^{0.24}_{0.19}$ & $0.22^{0.24}_{0.20}$ & $0.23^{0.25}_{0.20}$ & $0.23^{0.25}_{0.21}$ & $0.23^{0.26}_{0.21}$ & $0.24^{0.26}_{0.21}$ & $0.24^{0.25}_{0.22}$ & $0.24^{0.26}_{0.21}$ & $0.24^{0.26}_{0.22}$ & $0.24^{0.26}_{0.22}$ & $0.24^{0.27}_{0.22}$ & $0.24^{0.27}_{0.22}$ \\
    [2pt]
     & \textbf{w/o} & $0.21^{0.23}_{0.19}$ & $0.22^{0.25}_{0.20}$ & $0.24^{0.26}_{0.21}$ & \underline{$0.25$}$^{0.28}_{0.23}$ & \underline{$0.25$}$^{0.28}_{0.23}$ & \underline{$0.26$}$^{0.29}_{0.24}$ & \underline{$0.26$}$^{0.28}_{0.24}$ & \underline{$0.27$}$^{0.29}_{0.24}$ & \underline{$0.27$}$^{0.29}_{0.24}$ & \underline{$0.26$}$^{0.29}_{0.24}$ & $0.26^{0.29}_{0.24}$ & $0.25^{0.28}_{0.23}$ \\

    \midrule

    \multirow{2}{*}{\rotatebox[origin=c]{90}{\textbf{R-l}}} & \textbf{w/} & \underline{$0.20$}$^{0.22}_{0.17}$ & \underline{$0.21$}$^{0.23}_{0.18}$ & $0.22^{0.24}_{0.19}$ & $0.22^{0.24}_{0.20}$ & $0.22^{0.25}_{0.20}$ & $0.23^{0.25}_{0.20}$ & $0.23^{0.24}_{0.21}$ & $0.23^{0.25}_{0.20}$ & $0.23^{0.25}_{0.21}$ & $0.23^{0.25}_{0.20}$ & $0.23^{0.25}_{0.20}$ & $0.23^{0.25}_{0.21}$ \\
    [2pt]
     & \textbf{w/o} & $0.17^{0.20}_{0.15}$ & $0.19^{0.22}_{0.17}$ & $0.21^{0.23}_{0.18}$ & $0.22^{0.25}_{0.20}$ & $0.23^{0.25}_{0.21}$ & $0.24^{0.27}_{0.22}$ & $0.24^{0.26}_{0.23}$ & $0.24^{0.27}_{0.22}$ & $0.24^{0.27}_{0.22}$ & $0.24^{0.27}_{0.22}$ & $0.23^{0.26}_{0.21}$ & $0.23^{0.26}_{0.21}$ \\

    \bottomrule
    \end{tabular}
    \caption{95\% Confidence intervals (CIs) via per-query bootstrap for two base-sized model pairs: BERT-large (B-l), RoBERTa-large (R-l) on NanoBEIR datasets. Each column shows the average performance and CIs for the indicated number of remaining dimensions after truncation. A score is \underline{underlined} when it significantly outperforms its counterpart by two-sided paired t-test ($p < 0.05$).}
    \label{tab:ci-large}
\end{table}

\end{document}